%% file: preprint.tex
\documentclass[3p,times,preprint]{elsarticle}

\usepackage{amsmath}
\usepackage{amssymb}
\usepackage{algorithm}
\usepackage{booktabs}
\usepackage[dvipsnames]{xcolor}
\usepackage[hidelinks]{hyperref}
\hypersetup{pdfauthor=author}

\usepackage{lineno}
\usepackage{tabularx}
\usepackage{multirow}

\newcommand{\dashrule}[1][black]{%
  \color{#1}\rule[\dimexpr.5ex-.2pt]{3pt}{.4pt}\xleaders\hbox{\rule{3pt}{0pt}\rule[\dimexpr.5ex-.2pt]{3pt}{.4pt}}\hfill\kern0pt%
}

\begin{document}

\begin{frontmatter}




\title{Distributional Formal Semantics}


\author[a1]{Noortje J. Venhuizen\corref{corauth}}
\ead{noortjev@coli.uni-saarland.de}
\author[a2]{Petra Hendriks}
\author[a1]{Matthew W. Crocker}
\author[a1]{Harm Brouwer}

\cortext[corauth]{Corresponding author}

\address[a1]{Saarland University, Department of Language Science and Technology, 66123 Saarbr\"ucken, Germany}
\address[a2]{University of Groningen, Center for Language and Cognition Groningen (CLCG), P.O. Box 716, 9700AS Groningen, the Netherlands}

\begin{abstract}
  Natural language semantics has recently sought to combine the
  complementary strengths of formal and distributional approaches to
  meaning. More specifically, proposals have been put forward to augment
  formal semantic machinery with distributional meaning representations,
  thereby introducing the notion of semantic similarity into formal
  semantics, or to define distributional systems that aim to incorporate
  formal notions such as entailment and compositionality. However, given the
  fundamentally different `representational currency' underlying formal and
  distributional approaches---models of the world versus linguistic
  co-occurrence---their unification has proven extremely difficult. Here, we
  define a Distributional Formal Semantics that integrates distributionality
  into a formal semantic system on the level of formal models. This approach
  offers probabilistic, distributed meaning representations that are also
  inherently compositional, and that naturally capture fundamental semantic
  notions such as quantification and entailment. Furthermore, we show how
  the probabilistic nature of these representations allows for probabilistic
  inference, and how the information-theoretic notion of ``information''
  (measured in terms of Entropy and Surprisal) naturally follows from it.
  Finally, we illustrate how meaning representations can be derived
  incrementally from linguistic input using a recurrent neural network
  model, and how the resultant incremental semantic construction procedure
  intuitively captures key semantic phenomena, including negation,
  presupposition, and anaphoricity.
\end{abstract}


\end{frontmatter}




\input{introduction}

\input{framework}

\input{model}

\input{results}

\input{discussion}

\input{conclusion}

\appendix

\input{appendix}

\section*{Acknowledgements}

Funded by the Deutsche Forschungsgemeinschaft (DFG, German Research
Foundation) -- Project-ID 232722074 -- SFB 1102.




\bibliographystyle{elsarticle-num}
\bibliography{preprint}


\end{document}

%% file: introduction.tex
\section{Introduction}\label{sec:introduction}

Traditional formal approaches to natural language semantics capture the
meaning of linguistic expressions in terms of their logical interpretation
within abstract formal models. Central to these approaches---which range
from first-order predicate logic \cite{frege1879begriffsschrift} to
event-based \cite{davidson1967truth} and dynamic semantic approaches, such
as Discourse Representation Theory \cite{kamp1993discourse}---are the
notions of entailment and compositionality, which describe how meanings are
related to each other (entailment) and how they can be combined to form
complex meanings (compositionality). An alternative approach to natural
language semantics, which has recently gained much interest, is
distributional semantics. This approach characterizes the meaning of lexical
items in a usage-based manner; namely as distributional vectors that capture
the co-occurrences between words \cite{turney2010frequency,erk2012vector,
clark2015vector,lenci2018distributional}. The main advantage of such
approaches is that the distributional representations inherently encode
semantic similarity and relatedness between lexical items
\cite{landauer1997solution}, and that they can be derived empirically from
language data. It has, however, proven extremely difficult to incorporate
the traditional semantic notions of entailment and compositionality within
such a distributional semantic framework \cite{Boleda2016formal}. 

Indeed, while formal semantics focuses on proposition-level (sentence)
meanings and distributional semantics focuses on the level of words, there
has been considerable interest in bringing together the strengths of both
approaches within a single formalism. This has been attempted, for instance,
by defining a notion of composition on top of the distributional
representations, using vector operations \cite{mitchell2010composition}, or
by using more complex structures (e.g., matrices and tensors) in addition to
vectors to represent lexical expressions
\cite{baroni2010nouns,zanzotto2010estimating, coecke2011mathematical,
socher2012semantic,baroni2014frege,grefenstette2015concrete,rimell2016relpron}.
Although this has been shown to produce interesting results when applied to
adjective-noun modification \cite{vecchi2017spicy}, the approach has
difficulties in representing the meaning of complex, multi-argument
expressions (see \cite{blacoe2012comparison,lenci2018distributional} for
reviews). Conversely, it has been attempted to incorporate distributional
semantic representations within the logical forms that derive from formal
models of meaning
\cite{garrette2014formal,asher2016integrating,beltagy2016representing}. This
approach aims to exploit the complementary strengths of formal and
distributional semantics: formal mathematical machinery to account for
sentence-level inference combined with a distributional account for
lexical-level similarity and entailment \cite{chersoni2016towards}. While
approaches such as these do indeed go a long way in extending formal
semantics with a distributional component, they do so in a non-integrative
manner; that is, while distributional semantic representations can guide the
construction of logical form (e.g., \cite{asher2016integrating}), logical
inference remains part and parcel of the formal semantic system itself. This
is due to the fact that formal and distributional semantics operate on a
fundamentally different `representational currency': formal semantics
defines propositional meaning in terms of models of the world, whereas
distributional semantics defines lexical meaning in terms of linguistic
context.

An alternative approach from cognitive science, however, extends the
situation-state space framework from Golden and Rumelhart
\cite{Golden1993parallel} into a vector space model that defines meaning in
a distributed manner relative to states of the world
\cite{Frank2003modeling}. Here, we generalize this approach to reveal its
full logical capacity: we introduce a notion of distributionality into a
formal semantic system by defining meaning in terms of co-occurrence across
formal models. The resulting framework, called Distributional Formal
Semantics (DFS; as introduced in \cite{venhuizen2019framework}), defines a
meaning space for describing propositional-level meaning in terms of
co-occurrence across a set of logical models: individual models are treated
as observations, or cues, for determining the truth conditions of logical
expressions---analogous to how individual linguistic contexts are cues for
determining lexical meaning in distributional semantics. Based on a set of
propositions $\mathcal{P}$, we define a set of logical models
$\mathcal{M}_\mathcal{P}$ that together reflect the state of the world both
truth-conditionally and probabilistically (i.e., reflecting the
probabilistic structure of the world). The meaning of a proposition is
defined as a vector within the meaning space constituted by
$\mathcal{M}_\mathcal{P}$, which reflects its truth or falsehood relative to
each of the models in $\mathcal{M}_\mathcal{P}$. The meaning vectors capture
the (probabilistic) truth conditions of individual propositions indirectly
by defining the meaning of individual propositions in relation to all other
propositions; propositions that have related meanings will be true in many
of the same models, and hence have similar meaning vectors. In other words,
the meaning of a proposition is defined in terms of the propositions that it
co-occurs with---or, to paraphrase the distributional hypothesis formulated
by Firth \cite{Firth1957synopsis}: ``You shall know a \textit{proposition}
by the company it keeps''.

In what follows, we will introduce the DFS framework and show how its
meaning space offers compositional distributed representations that are
inherently probabilistic and inferential \citep[see
also][]{venhuizen2019framework}, and allow for capturing the
information-theoretic notions of Entropy and Surprisal
\citep[cf.][]{venhuizen2019expectation,venhuizen2019entropy}. We show how
propositional, as well as sub-propositional meanings are represented in the
DFS framework as points in the meaning space, and how the incremental
derivation of utterance meaning can be modeled as a trajectory through the
meaning space. To this end, we employ a recurrent network model that is
trained to map utterances onto their corresponding semantics on a
word-by-word basis. We show how each word induces a contextualized
transition from one point in meaning space to another, and critically extend
upon previous work by modeling the effect of various core semantic
phenomena---negation, presupposition, anaphoricity, and quantification---on
incremental semantic construction. Finally, we discuss how Distributional
Formal Semantics offers a powerful synergy between formal and distributional
approaches to meaning---which we argue to be complementary to the lexical
semantic knowledge captured by `traditional' distributional
approaches---that allows for modeling natural language semantics from
formal, empirical, and cognitive perspectives.

%% file: framework.tex
\section{A Framework for Distributional Formal Semantics}\label{sec:framework}

In Distributional Formal Semantics, meaning is defined relative to a
(finite) set of formal models $\mathcal{M_P}$, in which each model is
defined in terms of  a set of propositions $\mathcal{P}$. The propositions
in $\mathcal{P}$ may be simple (zero-ary predicates) or complex (predicates
with multiple arguments). The models in $\mathcal{M_P}$ are first-order
models that can be represented as the set of propositions $P \subseteq
\mathcal{P}$ that they satisfy. The set of models $\mathcal{M_P}$, which can
theoretically be viewed as a set of possible worlds in the tradition of
Carnap \cite{carnap1947meaning}, constitutes a meaning space relative to
which propositional meaning (as well as sub-propositional meaning; see
below) can be expressed. An example of the meaning space is shown in
Figure~\ref{fig:matrix}; here, the rows represent models as sets of (truth
values of) propositions and the columns represent propositional meaning
vectors. The meaning vector $\vec{v}(p)$ thus defines the meaning of
proposition $p \in \mathcal{P}$ in terms of the models that satisfy $p$;
i.e., $\vec{v}(p)$ is the vector that assigns $1$ to all $M\in\mathcal{M_P}$
such that $p$ is satisfied in $M$, and $0$ otherwise:
\begin{equation}
        \llbracket p \rrbracket^{\mathcal{M_P}} = \vec{v}(p) 
        \hspace{2em} \text{s.t. for all } i \in \{1,\ldots,|\mathcal{M_P}|\}: 
        \vec{v}_i(p) = 1 \text{ iff } M_i \vDash p 
\end{equation}

\begin{figure}
        \begin{center}
                $\hspace{1.5cm}\text{meaning vectors for propositions in } \mathcal{P}$\\
                $\hspace{1.5cm}\overbrace{\hspace{3.5cm}}$\\
                \rotatebox[origin=c]{90}{$\text{models } \mathcal{M_P}$\hspace{1em}}
                \rotatebox[origin=c]{90}{$\overbrace{\hspace{2cm}}$\hspace{1em}}
                \begin{tabular}{l | c | c | c | c | c}
                                & $p_1$ & $p_2$ & $p_3$ & \ldots & $p_n$\\
                        \hline
                        $M_1$  & 1 & 0 & 0 & \ldots & 1 \\
                        \hline
                        $M_2$  & 0 & 1 & 1 & \ldots & 1 \\
                        \hline
                        $M_3$  & 1 & 1 & 0 & \ldots & 0 \\
                        \hline
                        \ldots & . & . & . & \ldots & . \\
                        \hline
                        $M_m$  & 0 & 1 & 0 & \ldots & 0
                \end{tabular}
        \end{center}
        \caption{Example of a meaning space based on a set of models
        $\mathcal{M_P} = \{M_1 \ldots M_m\}$, with meaning vectors
        for the set of propositions $\mathcal{P} = \{p_1 \ldots p_n\}$.}
        \label{fig:matrix}
\end{figure}

Since propositional meaning is defined directly at the level of model
interpretation, propositions that are true in many of the same models will
obtain similar meaning vectors. That is, in the meaning space that derives
from $\mathcal{M_P}$, propositional meaning is defined in terms of
co-occurrence between propositions---rather than in terms of linguistic
co-occurrence, as in Distributional Semantics (e.g.,
\cite{landauer1997solution}).

In order for the meaning vectors to correctly capture propositional meaning
in terms of co-occurrence with other propositions, the meaning space should
be structured in such a way that it reflects the structure of the world,
both truth-conditionally and probabilistically. That is, in order to fully
encode entailment between two arbitrary propositions $p$ and $q$ ($p \vDash
q$), the set of models $\mathcal{M_P}$ should be truth-conditionally
complete, such that: $p \vDash q$ (for any $p,q \in \mathcal{P}$) iff all
models in $\mathcal{M_P}$ that satisfy $p$ also satisfy $q$, and $p \vDash
\neg q$ iff no model in $\mathcal{M_P}$ that satisfies $p$ also satisfies
$q$. The meaning space thus inherently encodes truth-conditional constraints
on propositional co-occurrence in terms of entailment; we will refer to such
constraints as \textit{hard} world knowledge constraints (see
Table~\ref{tab:constraints} in Section~\ref{sec:constructing} below for
examples of such hard world knowledge constraints). Moreover,
\textit{probabilistic} world knowledge is captured in the meaning space in
terms of the fraction of the models in $\mathcal{M_P}$ that satisfy a
certain proposition---or combination of propositions---from $\mathcal{P}$.
Hence, in order for the meaning space to reflect the structure of the world,
its constituent set of models $\mathcal{M_P}$ should capture the hard and
probabilistic world knowledge constraints dictated by the world. We here
present a top-down algorithm that induces a meaning space based on a
high-level specification of the world (but see the discussion in
Section~\ref{sec:datadriven} for an alternative, data-driven approach to
defining a meaning space for DFS). Accordingly, we describe the formal
properties of the resulting meaning space in terms of the compositionality
of the meaning vectors, their probabilistic and inferential properties, and
how it captures information-theoretic aspects of meaning (using the notions
of Entropy and Surprisal).

\subsection{Sampling the meaning space}\label{sec:sampling}

Given a set of propositions $\mathcal{P}$, the goal is to sample a set
of models $\mathcal{M_P}$ that reflects hard and probabilistic
constraints of the world (theoretically, there are $2^\mathcal{P}$ possible
models, but many of these are ruled out by hard constraints on propositional
co-occurrence). That is, each individual model $M\in\mathcal{M_P}$
should capture the truth-conditional constraints on co-occurrence between
the propositions in $\mathcal{P}$, and the full set of models
$\mathcal{M_P}$ should reflect any probabilistic constraints on
the (co-)occurrence of the propositions in $\mathcal{P}$. Following
\cite{venhuizen2019framework}, we employ an inference-driven,
non-deterministic sampling algorithm that stochastically generates a set of
models from a pre-defined set of propositions $\mathcal{P}$ and a set of
(hard and probabilistic) constraints on these propositions.

The models $M \in \mathcal{M_P}$ that constitute the meaning
space are defined as basic (first-order) models that can be represented as
tuples $\langle U_M,V_M \rangle$ consisting of a universe of entities $U_M$
and an interpretation function $V_M$ that assigns (sets of) entities to the
individual constants and predicates represented in $\mathcal{P}$. The set of
models $\mathcal{M_P}$ is sampled by incrementally constructing
individual models in a stochastic, inference-driven manner, based on the
closed-world assumption (i.e., the assumption that individual models
describe full states-of-affairs in terms of the truth and falsehood of the
propositions in $\mathcal{P}$). The universe $U_M$ of each $M \in
\mathcal{M_P}$ is defined as a set of entities $U_M = \{e_1 \dots
e_n\}$, based on the set of individual constants ($c_1 \ldots c_n$) that
derive from $\mathcal{P}$. The interpretation function $V_M$, in turn, is
initialized to map each constant onto a unique entity: $V_M(c_i) = e_i$.
Next, we incrementally construct the model $\langle U_M,V_M\rangle$ based on
the set of propositions $\mathcal{P}$, while taking into account the hard
and probabilistic world knowledge constraints. Critically, since $\langle
U_M,V_M\rangle$ only satisfies propositions that are assigned the truth
value ``true'', it does not distinguish between propositions that are
assigned the truth value ``false'', and those that are still undecided
during incremental sampling. Therefore, in parallel to $\langle
U_M,V_M\rangle$---which we call the \textit{Light World}\footnote{cf. The
Legend of Zelda: A Link to the Past (Nintendo, 1992).} model ($LM$) ---we
construct a \textit{Dark World} model $DM = \langle U_M,V'_M\rangle$ that
satisfies all propositions that are assigned the truth value ``false''
(relative to $LM$). Indeed, this parallel model construction allows us to
formulate not only the truth conditions of a particular expression using
constraints on the Light World, but also its ``falsehood conditions'', by
evaluating the complement of the constraints relative to the Dark World. For
instance, given a Light World constraint of the form $\forall x. R(x)$, any
incrementally constructed Light World that does not satisfy all propositions
of the form $R(x)$ will violate this constraint, even if the truth values of
these propositions are still undecided. By introducing a Dark World,
the falsehood of the constraint $\forall x. R(x)$ relative to the Light
World can be proven by finding an entity in the Dark World for which $R(x)$
holds (hence formalizing the following logical equivalence: $\neg \forall x.
R(x) \Leftrightarrow \exists x. \neg R(x)$). In other words, violation of
the constraint can be explicitly verified by checking whether the complement
of the constraint ($\exists x. R(x)$) is satisfied relative to the Dark
World. Table~\ref{tab:complement} shows the full set of rules that define
the complement of a formula $\varphi$.

\begin{table}
        \caption[Complement]{The complement $\overline{\varphi}$ of any
            well-formed formula $\varphi$ is defined based on the following
            rules:\footnotemark\label{fn:complement}}
        \label{tab:complement}
        \begin{center}
                \begin{tabularx}{.9\textwidth}{lcl X lcl X lcl}
                    \toprule
                        $\overline{\neg \varphi}$                   & $=$ & $\neg \overline{\varphi}$ & 
                        & $\overline{\varphi \oplus \psi}$         & $=$ & ($\overline{\varphi} \wedge \overline{\psi}) \vee (\neg \overline{\varphi} \wedge \neg \overline{\psi})$ & 
                        & $\overline{\exists x. \varphi}$           & $=$ & $\forall x. \overline{\varphi}$\\
                        $\overline{\varphi \wedge \psi}$            & $=$ & $\overline{\varphi} \vee \overline{\psi}$ & 
                        & $\overline{\varphi \rightarrow \psi}$     & $=$ & $\neg \overline{\varphi} \wedge \overline{\psi}$ & 
                        & $\overline{\forall x. \varphi}$           & $=$ & $\exists x. \overline{\varphi}$\\
                        $\overline{\varphi \vee \psi}$              & $=$ & $\overline{\varphi} \wedge \overline{\psi}$ & 
                        & $\overline{\varphi \leftrightarrow \psi}$ & $=$ & $(\neg \overline{\varphi} \wedge \overline{\psi}) \vee (\overline{\varphi} \wedge \neg \overline{\psi})$ & 
                        & $\overline{p}$                         & $=$ & $p$\\
                    \bottomrule
                \end{tabularx}
        \end{center}
\end{table}

\footnotetext{Note that the complement of the implication was reported
incorrectly in \cite{venhuizen2019framework}.}

Using the Light World ($LM$) and Dark World ($DM$) models, the sampling
algorithm incrementally constructs a model based on the set of propositions
$\mathcal{P}$, a set of probabilistic constraints on $\mathcal{P}$---i.e.,
the function $Pr(p,M)$ that returns a probability for proposition $p$ based
on model $M$ (see \ref{app:prob_constraints})---and a set of hard world
knowledge constraints $\mathcal{C}$. This sampling procedure is described in
Algorithm~\ref{alg:sampling} below (cf. \cite{venhuizen2019framework}). This
sampling algorithm uses the Light World and the Dark World to determine for
a randomly selected proposition $p$ whether its truth (or falsehood)
violates any world knowledge constraints.  If $p$ is consistent with respect
to both the Light World and the Dark World, it is determined
probabilistically whether the proposition is ``true'' or ``false'' (with
respect to the Light World). If, on the other hand, $p$ is only consistent
with respect to the Light World, it is inferred to be ``true'' with respect
to the Light World. Conversely, if $p$ is only consistent with respect to
the Dark World, it is inferred to be ``false'' with respect to the Light
World (i.e., $p$ is satisfied by the Dark World). Finally, if $p$ is neither
consistent with respect to the Light World nor with respect to the Dark
World, the current model is discarded. This state of affairs may arise from
an interaction between individual hard world knowledge constraints, e.g.,
when adding $p$ to the Light World violates one constraint and adding $p$ to
the Dark World satisfies the complement of another. This sampling procedure
is repeated until the truth or falsehood of each proposition
$p\in\mathcal{P}$ is determined. The final Light World model $\langle
U_M,LV_M \rangle$ will be added to the sampled set of models
$\mathcal{M_P}$. The full sampling algorithm described here is implemented
as part of the software package
\textsc{dfs-tools}.\footnote{\textsc{dfs-tools} is available as open source
under the Apache Licence, Version 2.0:
\url{https://github.com/hbrouwer/dfs-tools}}

\begin{algorithm}[t]
        \caption{Sampling algorithm for deriving a DFS meaning space based
        on a set of propositions $\mathcal{P}$, a set of probabilistic
        constraints $\{Pr(p,M)|p\in\mathcal{P}\}$, and a set of hard world
        knowledge constraints $\mathcal{C}$.}
        \label{alg:sampling}
\begin{enumerate}\small
                \item Let $\langle LM,DM\rangle$ be the tuple consisting of
                the initialized Light World model $LM = \langle
                U_M,LV_M\rangle$ and the initialized Dark World model $DM =
                \langle U_M,DV_M\rangle$;
                \item Verify whether the state of affairs $\langle
                LM,DM\rangle$ is consistent: $\langle LM,DM\rangle \vDash
                \mathcal{C}$ \textit{iff} for all constraints
                $c\in\mathcal{C}$: either $c$ is satisfied ($LM \vDash c$)
                or $c$ is not falsified ($DM \nvDash \overline{c}$). If  $\langle
                LM,DM\rangle \nvDash \mathcal{C}$, start again from step 1. 
                \item Select a random proposition $p\in\mathcal{P}$, such that $LM
                \nvDash p$ and $DM \nvDash p$;
                \item Let $LM_p$ be an extension of $LM$ such that
                $LM_p\vDash p$, and let $DM_p$ be an extension of $DM$ such
                that $DM_p\vDash p$;
                \item Check the consistency of the state of affairs in which
                $p$ is satisfied in the Light World ($\langle
                LM_p,DM\rangle$), and the state of affairs in which $p$ is
                satisfied in the Dark World ($\langle LM,DM_p\rangle$).
                \begin{itemize}
                        \item If $\langle LM_p,DM\rangle \vDash \mathcal{C}$
                        and $\langle LM,DM_p\rangle \vDash \mathcal{C}$, the
                        truth/falsehood of $p$ is determined
                        probabilistically: let $LM_p$ be the new Light World
                        $LM$ with probability $Pr(p,LM)$; otherwise, let
                        $DM_p$ be the new Dark World $DM$.
                        \item If $\langle LM_p,DM\rangle \vDash \mathcal{C}$
                        and $\langle LM,DM_p\rangle \nvDash \mathcal{C}$,
                        $p$ is inferred to be true: let $LM_p$ be the new
                        Light World $LM$.
                        \item If $\langle LM_p,DM\rangle \nvDash
                        \mathcal{C}$ and $\langle LM,DM_p\rangle \vDash
                        \mathcal{C}$, $p$ is inferred to be false: let $DM_p$
                        be the new Dark World $DM$.
                        \item If $\langle LM_p,DM\rangle \nvDash
                        \mathcal{C}$ and $\langle LM,DM_p\rangle \nvDash
                        \mathcal{C}$, the state of affairs is
                        inconsistent: start over from step 1.
                \end{itemize}
                \item Repeat from step 2 until all propositions in
                $\mathcal{P}$ are satisfied in either $LM$ or $DM$: 
                \item If $\langle LM,DM\rangle \vDash \mathcal{C}$, the
                final $LM$ is stored as a sampled model.
        \end{enumerate}
\end{algorithm}

\subsection{Compositionality in DFS}\label{sec:composition}

The meaning vectors from the DFS meaning space are fully compositional at
the propositional level. Since the meaning vector $\vec{v}(p)$ of a
proposition $p\in\mathcal{P}$ is defined in terms of whether or not it is
satisfied by each of the models $M\in\mathcal{M_P}$, its negation can be
defined straightforwardly as the complement of $\vec{v}(p)$, i.e., the
vector that assigns $1$ to all $M\in\mathcal{M_P}$ such that $p$ is not
satisfied in $M$, and $0$ otherwise:
\begin{equation}
        \llbracket \neg p \rrbracket^{\mathcal{M_P}} = 
        \vec{v}(\neg p) 
        \hspace{2em} \text{s.t. for all } i \in \{1,\ldots,|\mathcal{M_P}|\}: 
        \vec{v}_i(\neg p) = 1 \text{ iff } M_i \nvDash p 
\end{equation}

Similarly, the meaning of the conjunction $p \wedge q$, for $p,q\in\mathcal{P}$,
is defined as the meaning vector $\vec{v}(p\wedge q)$ that assigns $1$ to all
$M\in\mathcal{M_P}$ that satisfy both $p$ and $q$, and $0$ otherwise:
\begin{equation}
        \llbracket p \wedge q\rrbracket^{\mathcal{M_P}} = 
        \vec{v}(p\wedge q)  
        \hspace{2em} \text{s.t. for all } i \in \{1,\ldots,|\mathcal{M_P}|\}: 
        \vec{v}_i(p \wedge q) = 1 \text{ iff } M_i \vDash p \text{ and } M_i \vDash q
\end{equation}

Using the negation and conjunction operators, the meaning of any other
logical combination of propositions in the semantic space can be defined,
thus allowing for meaning vectors representing expressions of arbitrary
logical complexity. Moreover, these operations also allow for the definition
of quantification. Since $\mathcal{P}$ fully describes the set of
propositions expressed in $\mathcal{M_P}$, the (combined) universe
of $\mathcal{M_P}$ ($U_{\mathcal{M_P}} = \{u_1, \ldots,
u_n\}$) directly derives from $\mathcal{P}$. Universal and existential
quantification, then, can be formalized by replacing the quantifier variable
in the sub-formula with each of the entities in
$U_{\mathcal{M_P}}$, and combining them using conjunction and
disjunction, respectively:
\begin{equation}
        \llbracket \forall x \varphi \rrbracket^{\mathcal{M_P}} = 
        \vec{v}(\forall x \varphi)  
        \hspace{2em} \text{s.t. for all } i \in \{1,\ldots,|\mathcal{M_P}|\}: 
        \vec{v}_i(\forall x \varphi) = 1 
        \text{ iff } M_i \vDash \varphi[x \backslash u_1] \wedge \ldots \wedge \varphi[x \backslash u_n]  
\end{equation}
\begin{equation}
        \llbracket \exists x \varphi \rrbracket^{\mathcal{M_P}} = \vec{v}(\exists x \varphi)  
        \hspace{2em} \text{s.t. for all } i \in \{1,\ldots,|\mathcal{M_P}|\}: 
        \vec{v}_i(\exists x \varphi) = 1 
        \text{ iff } M_i \vDash \varphi[x \backslash u_1] \vee \ldots \vee \varphi[x \backslash u_n] 
\end{equation}
where $\varphi[x \backslash u]$ is defined as the formula $\varphi$ with
every instance of $x$ replaced by $u$. This formalization of quantification
in the meaning space constituted by $\mathcal{M_P}$ is
directly in line with traditional formalizations of quantification, in which
an assignment function is used to substitute variables for elements from the
model universe. The difference is that here variables are replaced by
elements from the combined universe of all models in
$\mathcal{M_P}$. As a result, quantification is strictly defined
with respect to the full set of models, rather than individual models: for
instance, $\forall x R(x)$ is only true in those models that satisfy $R(x)$
for the full set of entities in $U_{\mathcal{M_P}}$, not just for
those entities that occur in the current model. This ensures that
entailments are preserved across models, e.g., $\forall x. R(x)$ entails
$R(e)$ for all entities $e$ across all models in~$\mathcal{M_P}$.

\subsection{Probability and Inference in DFS}\label{sec:probability}

Propositional meaning vectors in the DFS meaning space are defined in terms
of the models that satisfy a proposition. As a result, the meaning vectors
are inherently probabilistic; that is, a proposition that is satisfied by a
large number of models has a high probability, and vice versa. Formally,
this means that the probability of an individual proposition is determined
by the fraction of models in $\mathcal{M_P}$ that satisfy this proposition
(following \cite{venhuizen2019framework}):
\begin{equation}
        \label{eq:prior_prob_bin}
        P(p) = \frac{|\{M \in \mathcal{M_P}~|~M \vDash p\}|}{|\mathcal{M_P}|}
        \hspace{2em} \text{for } p \in \mathcal{P}
\end{equation}
Given the compositional operations defined above, this means that the
probability of any logical combination of propositions can be defined; for
instance, the conjunctive probability of two propositions $p$ and $q$ is
defined as the fraction of models that satisfy both propositions $p$
and~$q$:
\begin{equation}
        \label{eq:conj_prob_bin}
        P(p \wedge q) = \frac{|\{M \in \mathcal{M_P}~|~M \vDash p 
        \text{ and } M \vDash q\}|}{|\mathcal{M_P}|} 
        \hspace{2em} \text{for } p,q \in \mathcal{P}
\end{equation}

These definitions define probabilities over propositional-level meanings
that are represented within the DFS meaning space as binary meaning
vectors---reflecting truth and falsehood with respect to the models in
$\mathcal{M_P}$. Critically, however, the meaning space constituted by
$\mathcal{M_P}$ is continuous, which means that intermediate points in
space---represented as real-valued vectors---constitute valid
(sub-propositional) meanings with respect to $\mathcal{M_P}$. Intuitively,
each component $i$ of a real-valued meaning vector $\vec{v}(a)$ can be
interpreted using fuzzy logic as describing the `degree of truth' of
sub-propositional meaning $a$ (e.g., representing a word or a sequence of
words) relative to model $M_i \in \mathcal{M_P}$. In other words,
real-valued vectors constitute meanings that cannot be directly expressed as
combinations of propositions, but rather reflect a degree of uncertainty
regarding the propositional-level meanings. More formally, based on the set
of models $\mathcal{M_P}$, we can define the vector space
$\mathbb{R}^{|\mathcal{M_P}|}$ that contains all real-valued vectors within
the dimensions of $\mathcal{M_P}$. This vector space consists of both binary
meaning vectors (constituting propositional meanings) and real-valued
vectors (constituting sub-propositional meanings), that all carry their own
probability. To capture the probabilistic properties of the vectors from the
vector space $\mathbb{R}^{|\mathcal{M_P}|}$, we extend the definitions for
prior and conjunctive probabilities defined above (see
Equations~\ref{eq:prior_prob_bin} and~\ref{eq:conj_prob_bin}) to account for
real-valued vectors by calculating the average value of their components
(following \cite{Frank2009connectionist}). That is, given that the
probability of propositions is defined in terms of the fraction of models
that satisfy a proposition, the probability of a sub-propositional
meaning---represented as a real-valued vector in which each component
represents a fuzzy `degree of truth'---can be defined as the sum of its
components divided by the number of models in $\mathcal{M_P}$:
\begin{equation}
        \label{eq:prior_prob_real}
        P(a) = \frac{1}{|\mathcal{M_P}|} \sum_i \vec{v}_i(a)
        \hspace{2em} \text{for } \vec{v}(a) \in \mathbb{R}^{|\mathcal{M_P}|}
\end{equation}

Given the fuzzy logic interpretation of the real-valued components of
meaning vectors, the conjunction of two (distinct) points in meaning space
$a$ and $b$ can standardly be defined using point-wise vector multiplication
\cite{Frank2009connectionist}. Importantly, since multiplying a real-valued
vector with itself does not necessarily result in the original vector (in
other words, the operation is a continuous t-norm that is non-idempotent),
we define the conjunctive probability of a point with itself as its prior
probability: $P(a \wedge a) = P(a)$. This ensures analogous behavior between
the probabilities associated with propositional meanings (i.e., binary
meaning vectors) and sub-propositional meanings (i.e., real-valued meaning
vectors). The conjunctive probability of two arbitrary points in meaning
space $a$ and $b$ (such that $a \neq b$) is then defined as follows:
\begin{equation}
        \label{eq:conj_prob_real}
        P(a \wedge b) = \frac{1}{|\mathcal{M_P}|} \sum_i \vec{v}_i(a)\vec{v}_i(b)
        \hspace{2em} \text{for } \vec{v}(a), \vec{v}(b) \in \mathbb{R}^{|\mathcal{M_P}|}
\end{equation}

The definitions of prior and conjunctive probability given in
Equations~\ref{eq:prior_prob_real} and~\ref{eq:conj_prob_real} extend the
prior and conjunctive probabilities of propositions
(Equations~\ref{eq:prior_prob_bin} and~\ref{eq:conj_prob_bin}) to
real-valued meaning vectors in the vector space
$\mathbb{R}^{|\mathcal{M_P}|}$. More specifically, in the case of
propositional meaning vectors, each model that satisfies a proposition $p$
(or combination thereof) contributes $\frac{1}{|\mathcal{M}|}$ of
probability mass to $P(p)$. Similarly, in the case of real-valued meaning
vectors, each model that satisfies sub-propositional meaning $a$ in a fuzzy
manner to degree $d$, contributes $\frac{d}{|\mathcal{M}|}$ of probability
mass to $P(a)$.

Given the definitions for prior and conjunctive probability, we can
calculate the conditional probability of any---propositional or
sub-propositional---meaning $a$ relative to any other---propositional or
sub-propositional---meaning $b$ in the meaning space:
\begin{equation}
        P(a\,|\,b) = \frac{P(a \wedge b)}{P(b)} 
\end{equation}

Hence, the meaning of an arbitrary point in meaning space is inherently
related---in terms of probabilistic co-occurrence---to any other point in
meaning space. As a result, the meaning vectors inherently encode logical
dependencies between propositions, and combinations thereof. We can
therefore employ the conditional probability between meaning vectors to
formally define entailment and probabilistic inference within the meaning
space (see also \cite{richardson2006markov}). That is, meaning vector $a$ is
entailed by meaning vector $b$ ($b\vDash a$) if the conditional probability
$P(a|b)$ equals $1$ (i.e., if $a$ and $b$ reflect propositional meanings,
this means that any model in $\mathcal{M_P}$ that satisfies $a$ also
satisfies $b$). In order to quantify probabilistic inference of $a$ given
$b$, the prior probability of $a$ needs to be taken into account: if the
conditional probability $P(a|b)$ is higher than the prior probability
$P(a)$, this means that $a$ is positively inferred from $b$; conversely, if
$P(a|b) < P(a)$, $a$ is negatively inferred from $b$. The following
inference score (see \cite{Frank2009connectionist}) quantifies this
probabilistic inference on a range from $+1$ to $-1$.
\begin{equation}\label{eq:inference}
\textit{inference}(a,b) = 
\begin{cases}
\frac{P(a\,|\,b)\,-\,P(a)}{1\,-\,P(a)} & \hspace{2em} \text{if } P(a\,|\,b) > P(a)\\
\frac{P(a\,|\,b)\,-\,P(a)}{P(a)} & \hspace{2em} \text{otherwise}\\
\end{cases}
\end{equation}

An inference score of $1$ indicates that meaning vector $a$ is
perfectly inferred from $b$ (i.e., $b$ entails $a$: $b \vDash a$), an
inference score of $-1$ indicates that the negation of $a$ is perfectly
inferred from $b$ ($b\vDash \neg a$), and any inference score in between
these extremes reflects either positive ($\textit{inference}(a,b) > 0$) or
negative probabilistic inference ($\textit{inference}(a,b) < 0$). It is
important to note that the inference score itself does not define a
probability: rather, it quantifies how much the truth of proposition $b$
increases (or decreases) the probability of proposition $a$, relative to its
prior probability. Hence, $\textit{inference}(a,b) = 0$ means that the
posterior probability $P(a|b)$ is equal to the prior probability $P(a)$;
i.e., knowing $b$ does not increase nor decrease the certainty in $a$. 

As described above, (propositional) meaning in DFS is defined in terms of
co-occurrence between propositions. Given that probabilities are defined for
all points in meaning space, the inference score can be employed to describe
the meaning of any real-valued vector $\vec{v}(a)$ relative to any other
vector, and, in particular, relative to any of the propositions
$p\in\mathcal{P}$ ($\textit{inference}(p,a)$, which quantifies how much
proposition $p$ is inferred from the meaning vector constituted by $a$).
Hence, all points in the meaning space---be it those constituted by a binary
meaning vector representing propositional meaning, or by a real-valued
vector representing sub-propositional meaning---are inherently probabilistic
as well as inherently related to each other, as can be quantified using the
inference score.

\subsection{Quantifying information in DFS}\label{sec:information}

The probabilistic nature of the meaning vectors in DFS also allows us to
characterize points in meaning space using information theory, as proposed
by Shannon \cite{shannon1948mathematical}. That is, information theory
defines the concept of ``information'' from a communicative perspective with
respect to the set of possible messages that can be used to determine a
particular state-of-affairs (e.g., there are six possible messages to
describe the outcome of a roll of a die). This is mathematically captured by
the notion of \textit{Entropy}, which quantifies the amount of uncertainty
in a given (communicative) state; i.e., states of high uncertainty (high
Entropy) on average require more messages (information, in \textit{bits}) to
be resolved than states of low uncertainty. In terms of the meaning space,
each point in space is defined relative to the set of models $\mathcal{M_P}$
that constitutes the meaning space. These models are specified as maximally
consistent sets of propositions, reflecting fully specified states of
affairs. Moreover, they can themselves be represented as vectors in meaning
space, namely as the vector $\vec{v}(M)$ that is defined as the conjunction
of all propositions $p\in\mathcal{P}$ that are satisfied in $M$ and the
negations of all propositions $p'\in\mathcal{P}$ that are not satisfied in
$M$. As a result, each point in meaning space inherently captures
uncertainty about which fully specified state of affairs (i.e., which
$M\in\mathcal{M_P}$) is the case. That is, the closer a point is to a fully
specified state of affairs, the less uncertainty it contains and hence the
lower its information value. To quantify this notion of Entropy,
\cite{venhuizen2019entropy} define a probability distribution over the set
of meaning vectors that identify (unique) models in $\mathcal{M_P}$, i.e.,
$\mathcal{V_{M_P}} = \{\vec{v}(M)~|~M \in \{M_i | M_i \in \mathcal{M_P}\}
\text{ and } \vec{v}_i(M) = 1 \textit{iff} M_i = M\}$. For a point in space
$a$, Entropy is defined as follows:
\begin{equation} 
        H(a) = - \sum_{\vec{v}(M) \in \mathcal{V_{M_P}}}
        P(\vec{v}(M)\,|\,a) \log P(\vec{v}(M)\,|\,a) 
\label{eq:entropy} 
\end{equation}
Following this definition, Entropy is zero if the point in meaning space
defined by $a$ identifies a unique model. If, on the other hand, all models
are equally likely (i.e., the probability distribution over all possible
models is uniform), Entropy will be maximal. 

Given this notion of Entropy, the logical and probabilistic properties of
the DFS meaning space can be directly linked to the information-theoretic
notion of uncertainty. Moreover, in the context of incremental language
processing, a change in Entropy, typically a reduction, is taken to induce
processing difficulty (\cite{hale2006uncertainty}; see
\cite{venhuizen2019entropy} for discussion). Similarly, processing effort
has been linked to the notion of Surprisal, which quantifies the expectancy
of words in context \cite{hale2001probabilistic,levy2008expectation}; the
less expected a word is in a given context, the higher its Surprisal, and
hence the higher its processing effort. Following
\cite{venhuizen2019expectation}, Surprisal can be defined within the meaning
space as reflecting the `expectedness' of a transition between two points in
meaning space, as triggered by an individual word or, more generally, a
message. That is, for a given message $m_{a,b}$ that triggers a transition
in meaning space from point $a$ to $b$, Surprisal is high if point $b$ is
unexpected given point $a$, and low otherwise:
\begin{equation}
        S(m_{a,b}) = -\log P(b\,|\,a)
        \label{eq:surprisal}
\end{equation}
In other words, the logarithm of the conditional probability between meaning
vectors $b$ and $a$ is inversely proportional to the processing effort
induced by the transition triggered by $m_{a,b}$. Indeed, this definition of
Surprisal captures the ``self-information'' \cite{shannon1948mathematical}
of a transition in meaning space, and Entropy reflects the average Surprisal
over all possible transitions from one point to the next within the meaning
space.

The information-theoretic notions of Entropy and Surprisal thus directly
derive from the probabilistic nature of the DFS meaning representations. As
such, they constitute a direct link between formal theories of entailment
and inference, and theories of incremental natural language processing. That
is, incremental processing in the meaning space entails navigating the
meaning space on a word-by-word basis, such that each word induces a
transition from one point to the next. These intermediate meaning states
represent the meaning that is constructed up to a given word, and derive
from the mapping of sentences onto (propositional) meaning vectors in the
meaning space. Although this mapping can in theory be formalized using a
semantic interpretation function, e.g., using set-theoretic machinery (see
section~\ref{sec:discussion} for a preliminary discussion), it can also be
approximated directly using a neural network model. The neural network
approach has the advantage that intermediate meaning states not only capture
probabilities deriving from the structure of the meaning space, but also
those deriving from the (probabilistic) structure of the language (see
\cite{venhuizen2019expectation}). Below, we show how such a model of
incremental semantic construction captures semantic phenomena such as
negation, quantification, presupposition and anaphoricity, and how these
phenomena affect the incremental processing dynamics of the model.

%% file: model.tex
\section{A Model of Incremental Semantic Construction}\label{sec:model}

Building a recurrent neural network model of incremental meaning
construction involves three core steps. First, we need to construct an
appropriate meaning space that allows for capturing the semantic phenomena
of interest, in this case negation, presupposition, anaphoricity, and
quantification. Secondly, we need to define a language $\mathcal{L}$ that
allows for describing (complex) situations in the meaning space, including
expressions pertaining to the relevant phenomena. Finally, we need to train
the neural network model to successfully map the utterances from
$\mathcal{L}$ onto their corresponding semantics on a word-by-word basis.
Below, we will discuss each of these steps in detail.

\subsection{Constructing a meaning space}\label{sec:constructing}

The first step is to construct set of models $\mathcal{M}_\mathcal{P}$ that
defines a meaning space, based on a confined set of propositions
$\mathcal{P}$. For the current model, we construct a set of propositions
using a set of predicates (\textit{enter(p,l)}, \textit{call(p,s)},
\textit{arrive(s)}, \textit{order(p,o)}, \textit{bring(s,o)},
\textit{pay(p)}), which are combined with one or more of the following
constants as arguments: persons ($p \in \{\textit{mike, will, elli,
nancy}\}$),\footnote{In the language $\mathcal{L}$ used in the current model
(described in Section~\ref{sec:language} below) ``will'' only occurs as a
proper name, not as an auxiliary verb.} places ($l \in \{\textit{bar,
restaurant}\}$), servers ($s \in \{\textit{barman, waiter}\}$), and orders
($o_{\textit{food}} \in \{\textit{fries,salad}\}$; $o_{\textit{drink}} \in
\{\textit{cola, water}\}$).  In addition, each of the constants is
associated with the one-place predicate ``\textit{referent(r)}'', which can
be interpreted as introducing the referent into the current discourse
context (cf. the universe of discourse referents, as formalized in Discourse
Representation Theory \cite{kamp1993discourse}). 

\begin{table}
    \caption{Hard constraints implemented in construction of the meaning
        space. Based on the set of propositions $\mathcal{P}$ (see text),
        the sampling algorithm samples a set of models
        $\mathcal{M}_\mathcal{P}$, such that each model $M \in
        \mathcal{M}_\mathcal{P}$ satisfies all of these constraints.}
    \label{tab:constraints}
    \small
    \begin{tabularx}{\linewidth}{c l X}
        \toprule
        \textbf{No.} & \textbf{Constraint} & \textbf{Description}\\
        \midrule
        1   & $\forall x \forall y \forall z(\textit{enter}(x,y) \wedge z\neq y \rightarrow \neg \textit{enter}(x,z))$ 
            &  A person can only enter a single place.\\
        2   & $\forall x \forall y_{f/d} \forall z_{f/d}(\textit{order}(x,y) \wedge z\neq y \rightarrow \neg \textit{order}(x,z))$ 
            &  A person can only order a single type of food/drink.\\
        3   & $\forall x (\neg (\textit{enter}(x,\textit{bar}) \wedge \textit{call}(x,\textit{waiter})))$ 
            & Waiter cannot be called in bar.\\
        4   & $\forall x (\neg (\textit{enter}(x,\textit{restaurant}) \wedge \textit{call}(x,\textit{bartender})))$ 
            & Bartender cannot be called in restaurant.\\
        5   & $\forall x (\neg (\textit{call}(x,\textit{waiter}) \wedge call(x,\textit{bartender})))$
            & A person can only call either waiter or bartender.\\
        6   & $\forall x \forall y ((\textit{enter}(x,y) \wedge \textit{pay}(x)) \rightarrow \exists z(\textit{order}(x,z)))$ 
            & Entering and paying implies that something is ordered.\\
        7   & $\forall x \forall y ((\textit{order}(x,y) \wedge \textit{pay}(x)) \rightarrow \exists z (\textit{bring}(z,y)))$
            & Ordering and paying implies that the order is brought.\\
        8  & $\forall P\forall x (\textit{P}(x) \rightarrow \textit{referent}(x))$ 
            & One-place predicates assert their argument.\\
        9  & $\forall R\forall x \forall y (\textit{R}(x,y) \rightarrow (\textit{referent}(x) \wedge \textit{referent}(y)))$ 
            & Two-place predicates assert both arguments.\\
        \bottomrule
    \end{tabularx}
\end{table}

Based on the resulting set of propositions $\mathcal{P}$ ($|\mathcal{P}| =
51$), a meaning space was constructed by sampling a set of 10K models
$\mathcal{M}_\mathcal{P}$ (using the sampling algorithm described in
Section~\ref{sec:sampling}, as implemented in \textsc{dfs-tools}), while
taking into account world knowledge in terms of hard and probabilistic
constraints on propositional co-occurrence.  Table~\ref{tab:constraints}
shows the hard constraints that are taken into account for sampling the
meaning space; i.e., each model $M \in \mathcal{M}_\mathcal{P}$ satisfies
all of these constraints. The first constraint ensures that each individual
model can be interpreted unambiguously in terms of the state of affairs it
describes; since no explicit temporal information is encoded in the
predicates, multiple `\textit{enter}' events for the same person would
obscure the event structure of individual models and therefore affect
entailment and inferencing. Constraints 2-5 pose general constraints on
event co-occurrence. Note that due to the probabilistic sampling strategy,
individual models satisfy only a subset of the propositions in
$\mathcal{P}$. Therefore, (non-)co-occurrences between individual
propositions need to be defined explicitly and do not necessarily follow
from combinations between other constraints (e.g., consider a model that
only contains \textit{call} and \textit{order} predicates for a particular
person: while it may satisfy constraints 1-4, it could violate constraint 5,
and hence be invalid). In order to make sure individual models capture the
canonical order of events, constraints 6-7 encode dependencies between event
occurrences. Finally, constraints 8-9 make sure that all arguments of a
predicate are instantiated as referents. This will allow us to model
presupposition and anaphoricity.

In addition to the hard constraints, the set of models
$\mathcal{M}_\mathcal{P}$ as a whole reflects the probabilistic structure
identified by a set of probabilistic constraints; see
\ref{app:prob_constraints} for the full set of probabilistic constraints
used for sampling the current meaning space. As follows from the sampling
algorithm defined in Section~\ref{sec:sampling}, the sampling probability of
a proposition $p$ is defined relative to a model $M$ ($Pr(p,M)$). The
intuition behind this is that the probability of $p$ may depend on the
propositions that are satisfied relative to the model $M$ that describes the
state-of-affairs constructed so far. For instance, the sampling probability
of person $p$ ordering food, $Pr(\textit{order}(p,o_{\textit{food}}),M)$, is
low in case $\textit{enter(p,bar)}$ is satisfied in the current model $M$,
while it is high in case $\textit{enter(p,restaurant)}$ is satisfied in $M$
(if none of these constraints is satisfied in $M$, the proposition
$\textit{order}(p,o_{\textit{food}})$ will be sampled using a base
probability; see \ref{app:prob_constraints}). During sampling, the
truth/falsehood of proposition $p$ is determined based on the probability
$Pr(p,M)$ only in case the truth/falsehood of $p$ cannot be inferred (based
on the Light World and Dark World constructed so far; see
Section~\ref{sec:sampling}). Hence, as the sampling probabilities may
interact with the hard world knowledge constraints described in
Table~\ref{tab:constraints}, the observed probabilities in the sampled
meaning space may only indirectly reflect these sampling probabilities (as
will be illustrated below).

\begin{figure}
    \centering
    \includegraphics[width=.9\textwidth]{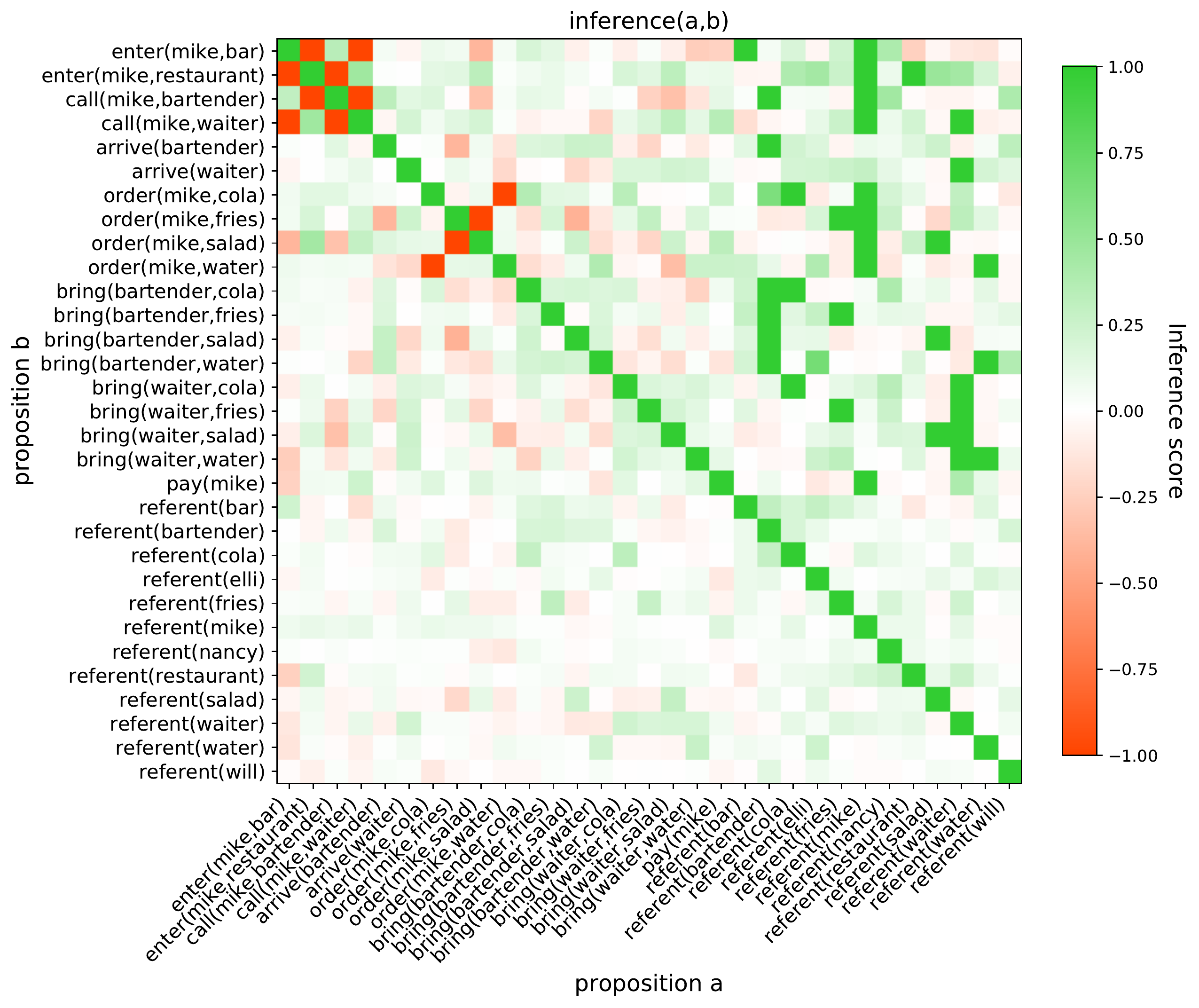}
    \caption{By-proposition inferences in the meaning space for a subset of
    the propositions: all \textit{referent} predicates together with
    all propositions pertaining to \textit{mike} and those pertaining to the
    servers \textit{waiter} and \textit{bartender}. Bright green values
    indicate maximal inference of proposition $a$ given proposition $b$
    (i.e., the entailment $b \vDash a$), white values indicate that there is
    no inference of proposition $a$ given $b$, and bright red values
    indicate maximal inference of $\neg a$ given $b$ (i.e., $b \vDash \neg
    a$); all intermediate values reflect positive (light green) or negative
    (light red) probabilistic inferences.}
    \label{fig:worldmap}
\end{figure}

To render it feasible to approximate the resultant meaning representations
using a neural network model, we reduced the set of 10K models to a
representative set of $150$ models, which captures the probabilistic and
truth-conditional structure of the world (using the model selection
algorithm described in \ref{app:selection}). To illustrate the richness of
the inferences contained in the resulting 150-dimensional meaning space,
Figure~\ref{fig:worldmap} shows the by-proposition inferences represented in
this space for a subset of the propositions.\footnote{Please refer to the
electronic version for color figures.} In this figure, bright green values
indicate maximal inference (i.e., entailment) of proposition $a$ given
proposition $b$, and bright red values indicate maximal inference of $\neg
a$ given proposition $b$. Every value in between reflects probabilistic
inference. The green diagonal shows that each proposition entails itself.
All other bright green and red values show (non-)co-occurrences between
propositions that derive from the hard world knowledge constraints; for
instance, \textit{order(mike,salad)} entails \textit{referent(mike)} and
$\neg$\textit{order(mike,fries)}. Furthermore, the probabilistic inferences
reflect the probabilistic world knowledge constraints described
in~\ref{app:prob_constraints}: for instance, the proposition
$a=\textit{order(mike,salad)}$ is negatively inferred given
$b=\textit{enter(mike,bar)}$ ($\textit{inference}(a,b) = -0.39$) and
positively inferred given $b'=\textit{enter(mike,restaurant)}$
($\textit{inference}(a,b') = 0.33$), hence following the pattern induced by
the probabilistic constraints (see constraints 5 and 6 in
Table~\ref{tab:prob_constraints}). The inference values shown in
Figure~\ref{fig:worldmap} thus quantify the structure of the meaning space
in terms of the hard and probabilistic co-occurrence constraints between
individual propositions, thereby providing a suitable means to compare
different meaning spaces. In fact, the model selection algorithm described
in \ref{app:selection} exploits exactly this information to derive a reduced
set of models that approximates the structure of the original meaning space.

\subsection{Mapping utterances onto semantics}\label{sec:language}

Based on the meaning space derived from the set of models
$\mathcal{M}_\mathcal{P}$, we define a language $\mathcal{L}$ based on a set
of words ($|W|=30$)\footnote{The negated auxiliary verb ``didn't'' is
considered a single word.} and a grammar that combines these words into
utterances (consisting of one or two sentences) that describe situations (in
terms of combinations of propositions) within the meaning space.
Table~\ref{tab:language} shows the utterances generated by the language
$\mathcal{L}$ and their associated semantics. Our grammar generates $278$
unique utterances that describe $270$ unique situations (due to the
restricted nature of the meaning space, in which there exists only a single
\textit{bar}/\textit{restaurant}, the definiteness of the determiner has no
effect on the meaning of assertive sentences, thus resulting in overlapping
semantics). The grammar generates three basic types of utterances: assertive
simple sentences (e.g., ``mike enters a restaurant''), negated simple
sentences (e.g., ``mike doesn't enter a restaurant') and assertive
two-sentence utterances (e.g., ``mike entered the bar he ordered
water'').\footnote{Two-sentence utterances are concatenated without
punctuation because in this small language it will have no effect on meaning
construction.} In addition, all two-sentence utterances also occur as
existentially quantified structures, which use the indefinite noun phrase
``someone'' (e.g., ``someone entered the restaurant he ordered salad''). The
semantics associated with assertive simple sentences are the individual
propositions described by these sentences, and asserted two-sentence
utterances describe conjunctions of propositions. The semantics of negated
simple sentences, in turn, is defined as a conjunction between the negated
proposition and the associated presupposed referents (i.e., those introduced
using a name or a definite determiner)---since each proposition entails all
its arguments as referents in the meaning space (see
Section~\ref{sec:constructing}), this is not explicitly encoded in the
semantics of assertive sentences. Finally, the utterances starting with
``someone'' obtain an existentially quantified version of their associated
semantics. Given the interpretation of existential quantification in DFS
(see Section~\ref{sec:composition}), this results in a disjunctive semantics
over all persons~($p \in \{\textit{mike, will, elli, nancy}\}$). 

\begin{table}
    \caption{Utterances in language $\mathcal{L}$ and their associated
        semantics. Each line contains two variants of the utterance (i.e.,
        asserted/negated or individual/quantified) with differential
        semantics, indicated using $a$ and $b$). Variables identify persons
        ($p \in \{\textit{mike, will, elli, nancy}\}$), places ($l \in
        \{\textit{bar, restaurant}\}$), servers ($s \in \{\textit{barman,
        waiter}\}$), and orders ($o \in \{\textit{cola, water, fries,
        salad}\}$). Utterances only describe possible situations in the
        meaning space and pronouns only occur with suitable antecedents
        (`$\exists x_{m/f}$' restricts quantifier scope to male/female
        entities based on the pronoun).}
    \label{tab:language}
    \small
    \begin{tabularx}{\linewidth}{ l l l }
        \toprule
        \textbf{Utterance$_{[a/b]}$} & \textbf{Semantics$_a$} & \textbf{Semantics$_b$}\\
        \midrule
        $p$ $[$entered/didn't enter$]$ a $l$    & \textit{enter(p,l)}   & $\neg$(\textit{enter(p,l)})  $\wedge$ \textit{referent(p)}\\
        $p$ $[$entered/didn't enter$]$ the $l$  & \textit{enter(p,l)}   & $\neg$(\textit{enter(p,l)}) $\wedge$ \textit{referent(p)} $\wedge$ \textit{referent(l)}\\
        $p$ $[$called/didn't call$]$ the $s$    & \textit{call(p,s)}    & $\neg$(\textit{call(p,s)}) $\wedge$ \textit{referent(p)} $\wedge$ \textit{referent(s)}\\
        $p$ $[$ordered/didn't order$]$ $o$      & \textit{order(p,o)}   & $\neg$(\textit{order(p,o)}) $\wedge$ \textit{referent(p)}\\
        $p$ $[$paid/didn't pay$]$               & \textit{pay(p)}       & $\neg$(\textit{pay(p)}) $\wedge$ \textit{referent(p)}\\
        the $s$ $[$arrived/didn't arrive$]$     & \textit{arrive(s)}    & $\neg$(\textit{arrive(s)}) $\wedge$ \textit{referent(s)}\\
        the $s$ $[$brought/didn't bring$]$ $o$  & \textit{bring(s,o)}   & $\neg$(\textit{bring(s,o)}) $\wedge$ \textit{referent(s)}\\
        $[p$/someone$]$ entered the $l$ $~$ he/she ordered $o$          & \textit{enter(p,l)} $\wedge$ \textit{order(p,o)}  & $\exists x_{m/f}(\textit{enter(x,l)} \wedge \textit{order(x,o)})$ \\
        $[p$/someone$]$ entered the $l$ $~$ he/she called the $s^\star$ & \textit{enter(p,l)} $\wedge$ \textit{call(p,s)}   & $\exists x_{m/f}(\textit{enter(x,l)} \wedge \textit{call(x,s)})$\\ 
        $[p$/someone$]$ called the $s$ $~~$ he/she ordered $o$          & \textit{call(p,s)} $\wedge$ \textit{order(p,o)}   & $\exists x_{m/f}(\textit{call(x,s)} \wedge \textit{order(x,o)})$\\
        $[p$/someone$]$ called the $s$ $~~$ he/she paid                 & \textit{call(p,s)} $\wedge$ \textit{pay(p)}       & $\exists x_{m/f}(\textit{call(x,s)} \wedge \textit{pay(x)})$\\
        $[p$/someone$]$ called the $s$ $~~$ he brought $o$              & \textit{call(p,s)} $\wedge$ \textit{bring(s,o)}   & $\exists x(\textit{call(x,s)} \wedge \textit{bring(s,o)})$\\
        $[p$/someone$]$ called the $s$ $~~$ he arrived                  & \textit{call(p,s)} $\wedge$ \textit{arrive(s)}    & $\exists x(\textit{call(x,s)} \wedge \textit{arrive(s)})$\\
        \bottomrule
    \end{tabularx}\\
    ~$^\star$ Only allowing possible combinations of Location--Server: \textit{bar--bartender}/\textit{restaurant--waiter}
\end{table}

Note that we here do not define a lexical semantics for individual words. We
use the compositional machinery from DFS to combine propositional meanings
into the meanings of entire utterances. To capture sub-propositional
meaning, we exploit the continuous nature of the meaning space. That is, the
meaning of a sub-propositional expression is a real-valued vector that
defines a point in the meaning space, which is positioned in between those
points that instantiate the propositional meanings that the expression
pertains to (e.g., the meaning of ``\textit{mike}'' will be expressed as the
meaning vector that is positioned in between the propositional meanings that
pertain to \textit{mike}; \textit{enter(mike,bar/restaurant)},
\textit{call(mike,bartender/waiter)}, etc.). In contrast to traditional
formal approaches, the DFS approach does not define an operation (such as
function composition) that simply combines the sub-propositional meanings of
two subsequent expressions. Rather, sequences of words $w_1 \ldots w_n$
define a trajectory $\langle \vec{v}_1,\ldots,\vec{v}_n \rangle$ through the
meaning space, where each $\vec{v}_i$ represents the (sub-propositional)
meaning induced by the sequence of words $w_1 \ldots w_i$; that is, each
word $w_i$ induces a meaning in the context of the meaning assigned to its
preceding words $w_1 \ldots w_{i-1}$. Sub-propositional meaning thus derives
from the incremental, context-dependent mapping from word sequences onto
(complex) propositional meanings. Below, we use a Simple Recurrent neural
Network (SRN), as proposed by Elman \cite{elman1990finding}, to approximate
this mapping (see also
\cite{venhuizen2019expectation,venhuizen2019framework}), and we show how the
structure of the meaning space and the language $\mathcal{L}$ combine in
incremental processing. In particular, we examine how the processing of
semantic phenomena such as negation, presupposition and anaphoricity emerges
from this behavior.

\subsection{Model specification}

\begin{figure}[t]
    \begin{center}
    \includegraphics[width=.6\linewidth]{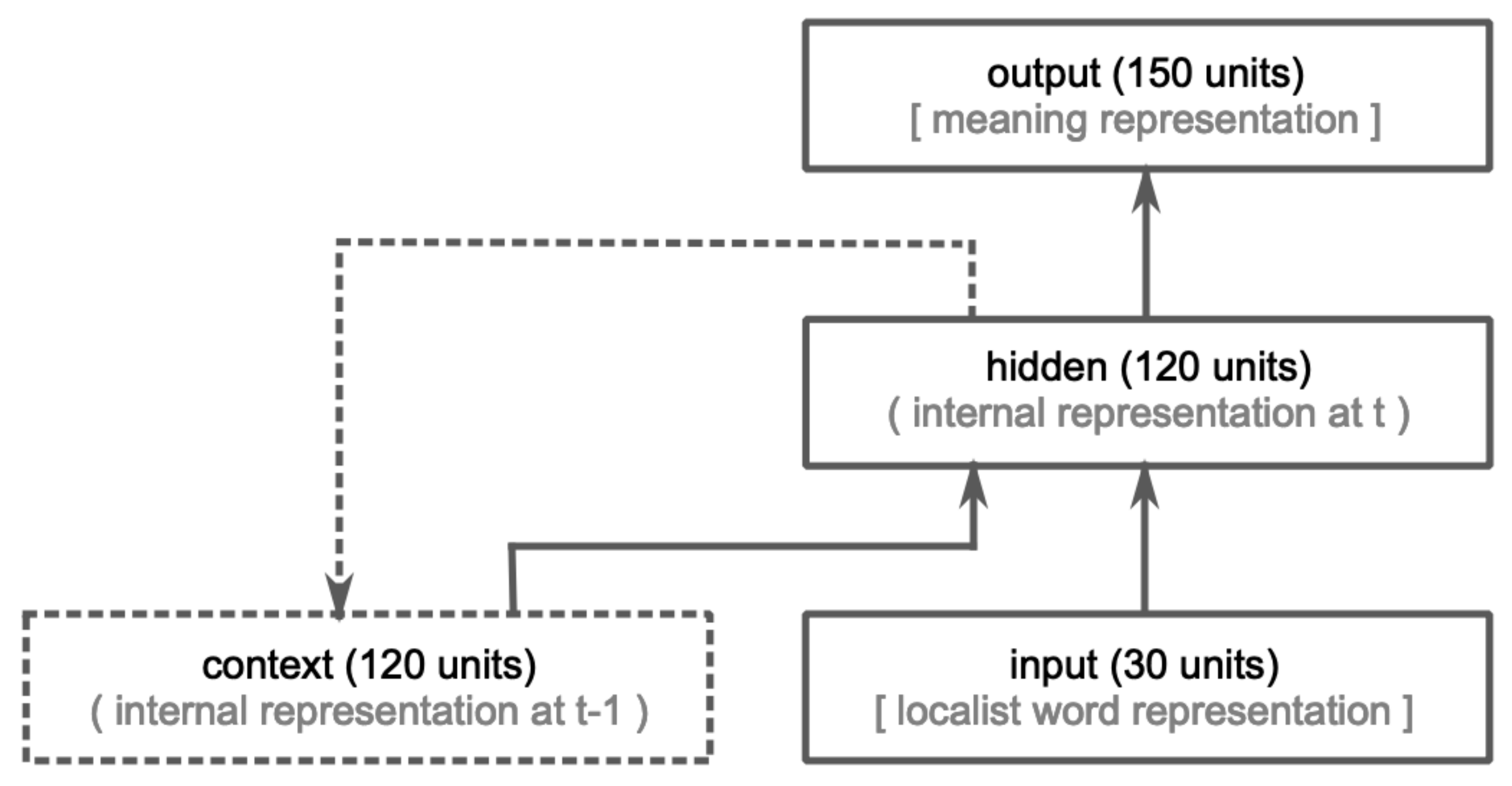}
            \caption{Simple Recurrent neural Network (SRN). Boxes represent
            groups of artificial neurons, and solid arrows between boxes
            represent full projections between the neurons in a
            projecting and a receiving group. The dashed lines indicate
            that the \textsc{context} layer receives a copy of the
            activation pattern at the \textsc{hidden} layer at the
            previous time-step. See text for details.}
            \label{figure:model}
    \end{center}
\end{figure}

In order to derive a mapping from word sequences onto DFS meaning vectors,
we train an SRN \cite{elman1990finding} to map sequences of words onto
meaning vectors that represent propositional-level (i.e., utterance-final)
meanings. The SRN consists of three groups of artificial logistic
dot-product neurons: an input layer (30 units), a hidden layer (120), and an
output layer (150) (see Figure \ref{figure:model}). Time in the model is
discrete, and at each processing time-step $t$, activation flows from the
input through the hidden layer to the output layer. In addition to the
activation pattern at the input layer, the hidden layer also receives its
own activation pattern at time-step $t-1$ as input (effectuated through an
additional context layer, which receives a copy of the activation pattern at
the hidden layer prior to feedforward propagation). The hidden and the
output layers both receive input from a bias unit (omitted in Figure
\ref{figure:model}). We trained the model using bounded gradient descent
\cite{Rohde2002connectionist} to map sequences of words onto a meaning
vector representing the meaning of that utterance relative to the set of
models $\mathcal{M}_\mathcal{P}$. More specifically, at each time step~$t$
during training, the model is presented with an individual localist word
meaning representation at its input layer (a vector with a single hot
bit)\footnote{We use localist representations in order not to presuppose any
word-internal structure. For more psychologically plausible representations,
it is possible to employ representational schemes that encode phonetic,
orthographic and/or semantic overlap between words.}, reflecting the current
word in the unfolding utterance, and an utterance-final meaning vector at
its output layer. At each time step $t$, the model thus effectively combines
a word meaning representation (from the input layer) with an abstract
representation of its context (reflected in the context layer) into a
meaning vector in the DFS meaning space (at the output layer). Since
individual words in context may map onto different utterance-final meanings
with varying frequencies, at each processing step $t$ the model will produce
a vector at its output layer that represents an abstraction over all
possible utterance-final meanings. Critically, since the model is trained to
map words onto meaning vectors, this output vector itself constitutes a
point in meaning space (see Section~\ref{sec:navigation} below).

Prior to training, the model's weights were randomly initialized within the
range of $(-.5,+.5)$. Each training item consisted of an utterance (a
sequence of words represented by localist representations) and a meaning
vector representing the utterance-final meaning. For each training item,
error was backpropagated after each word, using a zero error radius of
$0.05$, meaning that no error was backpropagated if the error on a unit fell
within this radius. Weight gradients were accumulated over epochs consisting
of all training items. At the end of each epoch, weights were updated using
a learning rate coefficient of $0.2$ and a momentum coefficient of $0.9$.
Training lasted for $10000$ epochs, after which the mean squared error was
$0.79$. The overall performance of the model was assessed by calculating
the cosine similarity between each utterance-final output vector and each
target vector for all utterances in the training data. After training, all
output vectors had the highest cosine similarity to their own target (mean
$=.99$; sd $=.01$), indicating that the model successfully learned to map
utterances onto their corresponding semantics. We moreover computed how well
the intended target could be inferred from the output of the model:
$\textit{inference}(\vec{v}_{target},\vec{v}_{output}$). The average
inference score over the entire training set was $0.88$, {which means that
after processing an utterance, the model almost perfectly infers the
intended meaning of the utterance.

\subsection{Semantic construction as meaning space navigation}\label{sec:navigation}

\begin{figure}[t]
        \centering
        \includegraphics[width=.75\linewidth]{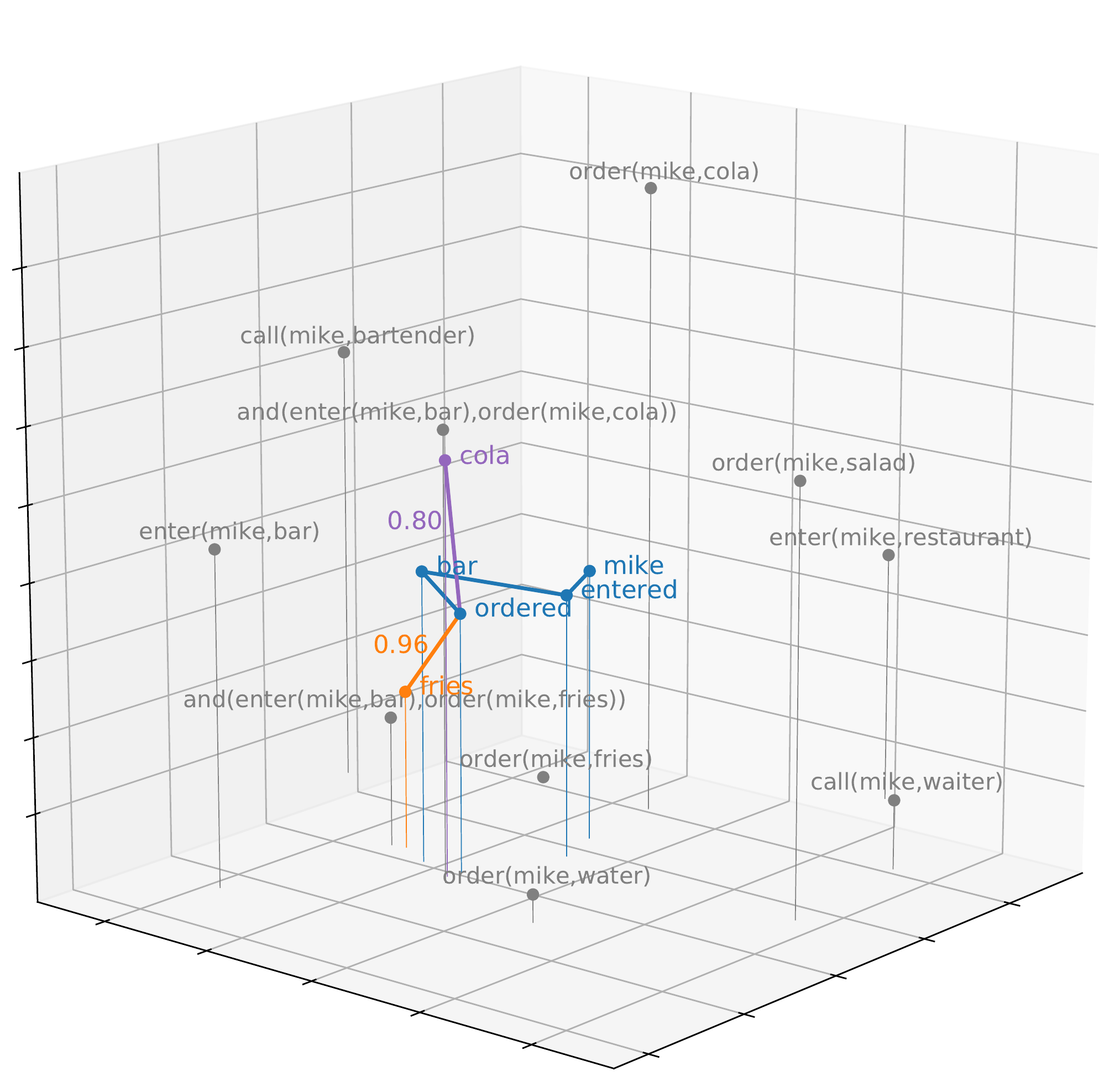}
                \caption{Visualization of the meaning space into three
                dimensions (using multidimensional scaling---MDS; see also
                Footnote~\ref{footnote7}) for a subset of the atomic
                propositions (those pertaining to \textit{mike}, excluding
                \textit{referent}), and the conjunctive meanings
                $\textit{enter(mike,bar)}\wedge\textit{order(mike,cola)}$
                and
                $\textit{enter(mike,bar)}\wedge\textit{order(mike,fries)}$.
                Grey points represent (simple or complex)
                propositional meaning vectors. Coloured points and lines
                show the word-by-word navigational trajectory of the model
                for the word sequence ``mike entered [the] bar [he]
                ordered'' (words in brackets are left out) with the
                continuations ``cola'' and ``fries''. The numbers
                represent Surprisal values for the utterance-final words.}
                \label{figure:cube}
\end{figure}

The model is trained to map each word in an utterance to the utterance-final
meaning. As a single word sequence may occur in multiple items with
different utterance-final meanings, the model learns to navigate the meaning
space on a word-by-word basis. Figure~\ref{figure:cube} provides a
visualization of this navigation process. This figure is a three-dimensional
representation of the $150$-dimensional meaning space (for a subset of the
atomic propositions and conjunctive meanings, derived using multidimensional
scaling (MDS).\footnote{\label{footnote7}Multidimensional scaling from $150$
into $3$ dimensions necessarily results in a significant loss of
information. Therefore, distances between points in the meaning space shown
in Figure~\ref{figure:cube} should be interpreted with care. The grey points
in this space correspond to propositional meaning vectors.} As this figure
illustrates, meaning in DFS is defined in terms of co-occurrence;
propositions that co-occur frequently in $\mathcal{M}_\mathcal{P}$ (e.g.,
\textit{enter(mike,restaurant)} and \textit{call(mike,waiter)}) are
positioned relatively close to each other (remember that the \textit{waiter}
can only be called in the \textit{restaurant}). The coloured points show the
model's word-by-word output for the word sequence ``mike entered the bar he
ordered'', with the continuations ``cola'' and ``fries''. The navigational
trajectory (indicated by the thick solid lines) illustrates how the model
assigns intermediate points in meaning space to sub-propositional
expressions, and approximates the utterance-final meaning at the final word.

Crucially, as shown in detail by \cite{venhuizen2019expectation}, this
trajectory is determined by the probabilistic structure of the meaning space
(``world knowledge'') as well as the utterances on which the model was
trained (``linguistic experience''). That is, at the word ``mike'', the
model navigates to a point in meaning space that is in between the meanings
of the propositions pertaining to \textit{mike}. Each consecutive word then
results in a transition to a new point in space that best approximates the
meaning up to that word. At ``bar'', the model positions itself close to the
meaning vector representing \textit{enter(mike,bar)}, but does not fully
commit to that meaning since its linguistic experience dictates that more
input may come, which would result in a conjunctive utterance-final
semantics. The effect of world knowledge, in turn, becomes clear at the
final words. While the model was exposed to the utterances ``mike entered
the bar he ordered cola'' and ``mike entered the bar he ordered fries''
equally often, the vector for ``mike entered the bar he ordered'' is closer
to $\textit{enter(mike,bar)}\wedge\textit{order(mike,cola)}$ than to
$\textit{enter(mike,bar)}\wedge\textit{order(mike,fries)}$, because the
former is more probable in the model's knowledge of the world. This
expectation is reflected in the Surprisal values (see
Equation~\ref{eq:surprisal}) associated with the utterance-final words:
``fries'' is more surprising ($S(\textit{fries}) = .96$) than ``cola''
($S(\textit{cola}) = .80$). For an elaborate investigation of the influence
of world knowledge and linguistic experience on meaning space navigation and
Surprisal, see \cite{venhuizen2019expectation}.

%% file: results.tex
\section{Inference during Semantic Construction}\label{sec:results}

The model maps utterances onto their corresponding semantics by navigating
the meaning space on a word-by-word basis. Each word triggers a transition
in meaning space from a point representing the meaning prior to encountering
that word to a point that integrates it into the interpretation.  Crucially,
each transition reflects both the linguistic experience of the model, as
well as by the world knowledge captured by the structure of the meaning
space, and each point directly allows for probabilistic inferences about what
is `understood'. That is, we can use the inference score described in
Section~\ref{sec:probability} (see Equation~\ref{eq:inference}) to quantify
how much a proposition $p$ is inferred after processing word $w_t$:
$\textit{inference}(\vec{v}(p),\vec{v}(w_t))$.

\begin{figure}
        \centering
        \includegraphics[width=.9\textwidth]{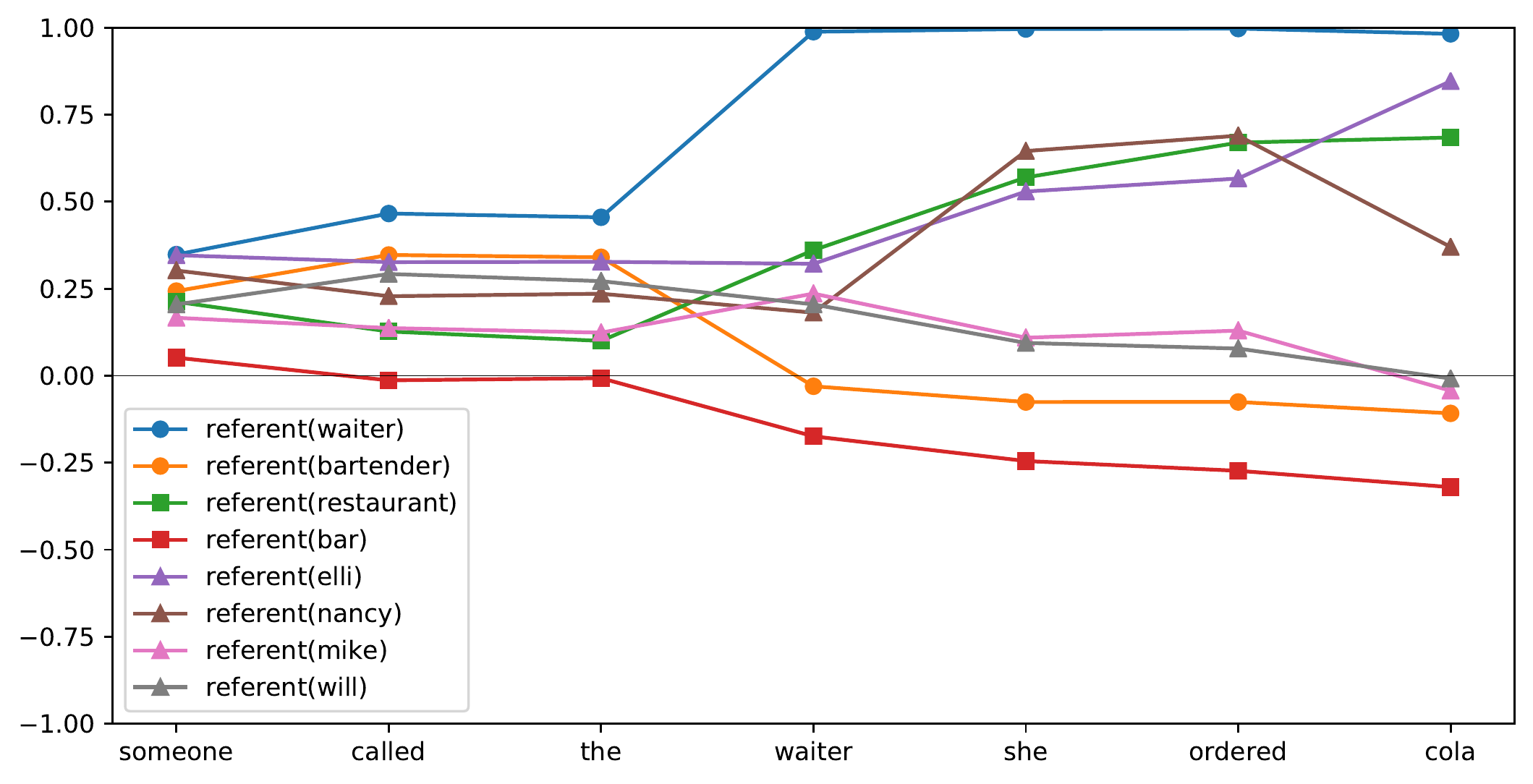}
        \caption{Word-by-word inferences for the utterance ``someone
        called the waiter she ordered cola'' for a subset of the
        \textit{referent} predicates. Inferences are calculated based on the
        word-by-word output of the model ($\vec{v}(w_t)$) and each of the
        propositions $p$, as follows:
        $\textit{inference}(\vec{v}(p),\vec{v}(w_t))$. Identical line
        markers reflect arguments of the same type
        (person/location/server).}
        \label{fig:wordbyword}
\end{figure}

Figure~\ref{fig:wordbyword} shows the word-by-word inferences for the
utterance ``someone called the waiter she ordered cola'' for a selection of
the \textit{referent} propositions. This figure shows that at each word in
the sentence, the model adjusts its inferences based on its linguistic
experience and the structure of the meaning space. For instance, at the word
``waiter'', the model perfectly infers that \textit{referent(waiter)} is the
case (blue line), and at the same time reduces its inference for
\textit{referent(bartender)} (orange line)---note that the hard world
knowledge constraints do not exclude models that satisfy both
\textit{referent(waiter)} and \textit{referent(bartender)}. Furthermore, at
the sentence-final word ``cola'', the model strongly infers
\textit{referent(restaurant)} (green line), while showing a negative
inference for  \textit{referent(bar)} (red line), despite the fact that
neither of these referents were mentioned in the sentence. This is an effect
of the structure of the meaning space, in which \textit{referent(waiter)}
often co-occurs with \textit{referent(restaurant)}, due to the hard world
knowledge constraints (see Table~\ref{tab:constraints}). Finally, this
figure shows that although the interpretation of the existentially
quantified expression ``someone'' remains underspecified throughout the
utterance, both the linguistic input (i.e., ``she'') and the structure of
the world (i.e., \textit{elli} is more likely to order \textit{cola} than
\textit{nancy}) guide the model toward an utterance-final interpretation in
which \textit{referent(elli)} (purple line) is inferred to a stronger degree
than \textit{referent(nancy)} (brown line).

Indeed, the abilitiy to quantify the degree to which each given proposition
is inferred at each point in the meaning space offers a powerful means to
examine the dynamics of incremental semantic construction. In what follows,
we will therefore harness probabilistic inference to investigate the
construction dynamics of four key semantic phenomena: negation,
presupposition, anaphoricity and quantification.

\subsection{Negation} 

\begin{figure}
        \centering
        \includegraphics[width=\textwidth]{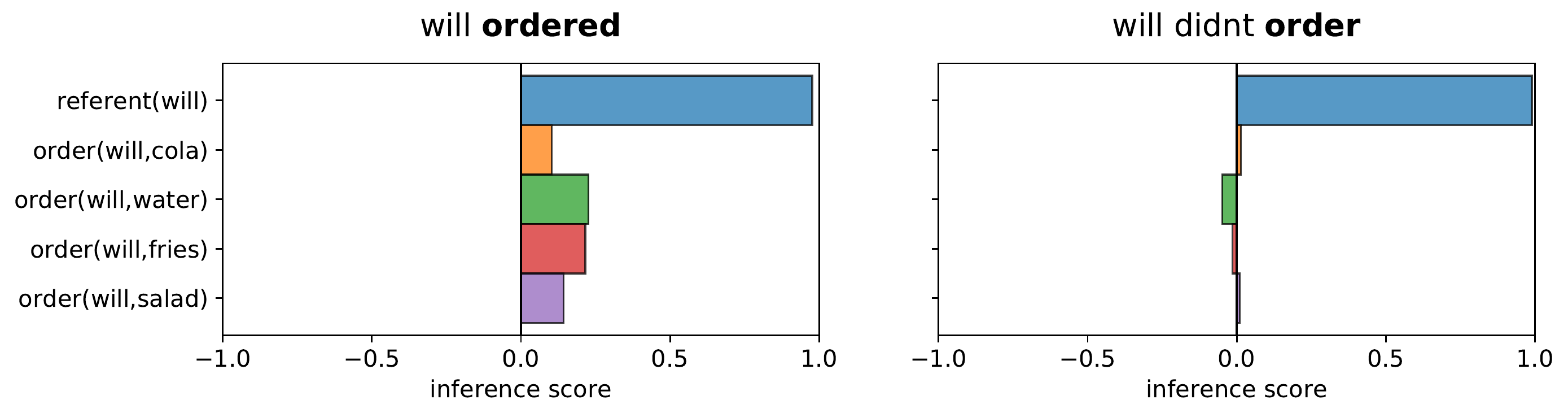}
        \caption{Processing of negation. The barplots show the inference
         scores for a relevant set of propositions given the model's output
         for the word sequences ``will \textbf{ordered}'' (left) and ``will didnt
         \textbf{order}'' (right)---critical words are shown in boldface.}
        \label{fig:negation}
\end{figure}

As described above, the meaning space inherently captures negation in terms
of the models that do not satisfy (combinations of) propositions. This
implies that each individual model in $\mathcal{M}_\mathcal{P}$ is
interpreted as a full description of a state of affairs (rather than a
partial observation, as in \cite{Frank2003modeling}), in which truth and
falsehood directly reflect the state of the world. The negation semantics
resulting from this meaning space were employed as the target semantics for
the negated sentences presented to the model (see Table~\ref{tab:language}).
Figure~\ref{fig:negation} shows the contrast between an assertive (left) and
a negated word sequence (right) for a selected set of inferences. 

Both word sequences can be continued by introducing one of the possible
orders (\textit{cola, fries, salad, water}). At the word ``ordered'' in the
utterance ``will ordered'' (left), the model has inferred that
\textit{referent(will)} is the case (blue bar). Moreover, the expected
continuations are reflected by positive inferences for each of the
propositions describing \textit{will} ordering something (note that the
personal preference for \textit{water} over \textit{cola} is also reflected
in these expectations). By contrast, the negated word sequence ``will didnt
order'' does not induce any of these probabilistic inferences; that is,
although \textit{referent(will)} is still inferred to be the case (see
Section~\ref{sec:presupposition} below), there are no (positive or negative)
inferences about any of the \textit{order} propositions (small deviations
from $0$ are interpreted as noise resulting from the model's navigation
through meaning space). This can be explained by the fact that the
\textit{order} predicates are partially mutually exclusive (\textit{cola}
excludes \textit{water}, and \textit{salad} excludes \textit{fries}), and
therefore all models $M\in\mathcal{M}_\mathcal{P}$ satisfy the negation of
at least two of the \textit{order} predicates. As a result, all models
satisfy ``will didnt order', which means that all inferences approximate $0$
(as the score \textit{inference(a,b)} quantifies the inference of $a$ above
and beyond its prior). Hence, negation---as part of the language presented
to the model during training, as well as deriving from the structure of the
meaning space---directly affects incremental semantic construction in terms
of the online expectations represented in the model.

\subsection{Presupposition}\label{sec:presupposition}

In order to capture (existential) presuppositions, the meaning space was
constructed in such a way that all individual constants instantiated in a
single model $M\in \mathcal{M}_\mathcal{P}$ were explicitly introduced using
the \textit{referent} predicate (similar to the instantiation of a universe
of discourse referents in Discourse Representation Theory
\cite{kamp1993discourse}; see also \cite{venhuizen2018discourse}). In the
mapping of utterances onto semantics, these `existential' predicates were
explicitly incorporated to account for the imbalance between definite and
indefinite noun phrases with regard to negation: whereas definite
descriptions trigger existence presuppositions that `project' from the scope
of the negation (in line with \cite{strawson1950referring}), indefinite
descriptions generally do not trigger such presuppositions.\footnote{We here
do not consider the use of `specific indefinites', which do elicit such
presuppositions in certain contexts (see, e.g.,
\cite{kasher1976semantics,geurts2010specific}).}

\begin{figure}
        \centering
        \includegraphics[width=\textwidth]{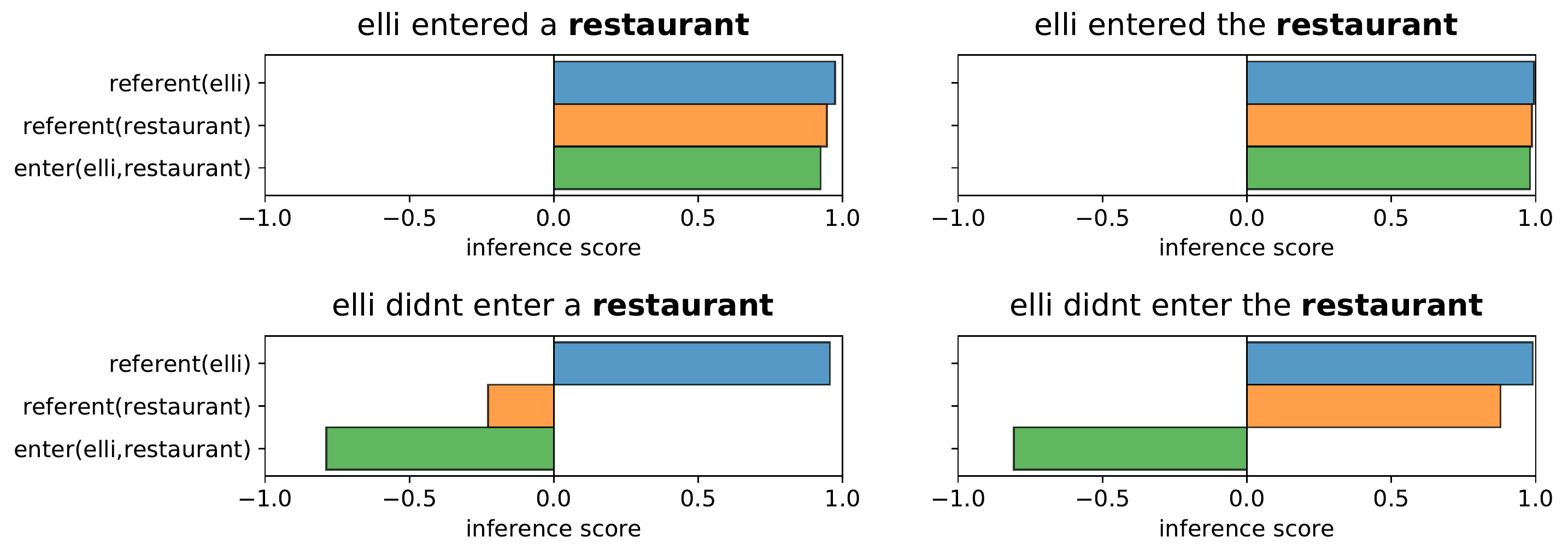}
        \caption{Processing of presupposition. The barplots show the inference
        scores for a relevant set of propositions given the model's output for
        the utterances ``elli entered/didnt enter a \textbf{restaurant}''
        (top/bottom left) and ``elli entered/didnt enter the
        \textbf{restaurant}'' (top/bottom right)---critical words are shown in
        boldface.}
        \label{fig:presupposition}
\end{figure}

Figure~\ref{fig:presupposition} shows how these manipulations manifest in
the online comprehension behavior of the model. The two panels at the top
show that as part of an assertive sentence, definite and indefinite noun
phrases result in the same inferences: the sentences ``elli entered a
restaurant/ elli entered the restaurant'' both result in strong positive
inferences (entailments) for \textit{referent(elli)} (blue),
\textit{referent(restaurant)} (orange) and \textit{enter(elli,restaurant)}
(green). When embedded in a negated sentence, however, the inferences start
to diverge. After processing the sentence ``elli didnt enter a restaurant''
(bottom left panel), the model infers that \textit{referent(elli)} is the
case and \textit{enter(elli,restaurant)} is not the case. Critically,
\textit{referent(restaurant)} is also negatively inferred, indicating that
the model finds itself in a state in meaning space in which
\textit{restaurant} is unlikely to be a referent (since
$\neg\textit{enter(elli,restaurant)}$ can co-occur with other propositions
that induce \textit{referent(restaurant)}, this inference is not maximally
negative). The sentence ``elli didnt enter the restaurant'' (bottom right
panel), on the other hand, does show a strong positive inference for
\textit{referent(restaurant)}, while at the same time inferring
$\neg\textit{enter(elli,restaurant)}$. This shows that the model is able to
interpret presuppositions in context and to adjust its inferences
accordingly.

\subsection{Anaphoricity}

The model was not explicitly trained to resolve the anaphoric dependencies
of anaphoric expressions. Instead, anaphoric expressions (pronouns) were
presented to the model as part of two-sentence utterances, which were
associated with the appropriate (discourse-final) semantics. Crucially,
pronouns were used to refer to both persons and servers (the latter always
using the pronoun ``he''), but their anaphoric antecedent was always
disambiguated by either the prior context or the following sentence.

\begin{figure}
        \centering
        \includegraphics[width=\textwidth]{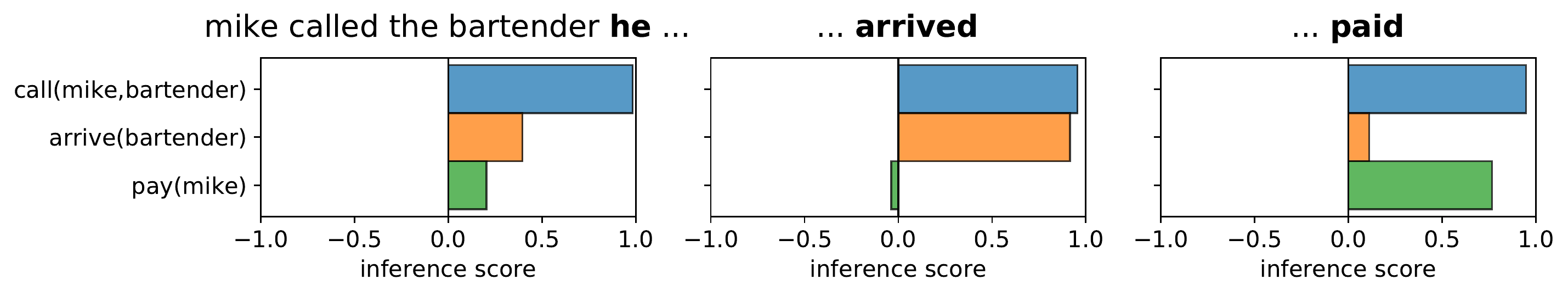}
        \caption{Processing of anaphora. The barplots show the inference
        scores for a relevant set of propositions given the model's output
        for the word sequence ``mike called the bartender \textbf{he}'' (left) and
        the continuations ``mike called the bartender he \textbf{arrived}''
        (middle) and ``mike called the bartender he
        \textbf{paid}'' (right)---critical words are shown in boldface.}
        \label{fig:anaphoricity}
\end{figure}

Figure~\ref{fig:anaphoricity} shows an example of referential ambiguity of
the pronoun. When processing the word ``he'' after the sentence ``mike
called the bartender'' (left panel), the model shows moderately positive
inferences for the propositions associated with possible continuations: both
\textit{arrive(bartender)} (orange) and \textit{pay(mike)} (green) are
positively inferred. Despite the fact that both continuations are equally
likely based on the model's linguistic experience,
\textit{arrive(bartender)} is inferred to a slightly larger degree, which
can be ascribed to a difference in conditional probabilities in the meaning
space ($P(\textit{arrive(bartender)} ~|~ \textit{call(mike,bartender)}) =
.62$; $P(\textit{pay(mike)} ~|~ \textit{call(mike,bartender)}) =
.46$).\footnote{The model selection procedure in some cases results in
(slight) probabilistic inferences that were not explicitly part of the
probabilistic constraints of the world; in particular, reducing the number
of models in $\mathcal{M}_\mathcal{P}$ may amplify co-occurrence
probabilities between propositions.} When the model encounters the
utterance-final word ``arrived'' (middle panel), the meaning is
disambiguated to the vector representing $\textit{call(mike,bartender)}
\wedge \textit{arrive(bartender)}$ (because \textit{arrive(mike)} is not a
valid proposition in the meaning space). Instead, when the utterance is
continued with ``paid'' (right panel), the model infers that
\textit{pay(mike)} is the case (again, \textit{pay(bartender)} is not part
of the meaning space), while still maintaining a slightly positive belief
for \textit{arrive(bartender)}. 

Hence, after processing the ambiguous pronoun ``he'', the model shows
positive inferences for both possible continuations, with a slight
preference for one over the other. In other words, the inferences reflect
the expectations of the model in terms of the possible continuations. We can
use the notion of Surprisal (see Equation~\ref{eq:surprisal}) to quantify
these expectations. Indeed, the Surprisal estimates reflect that the
continuation ``arrived'' is more expected than ``paid'', because the former
is less surprising ($S(\textit{arrived}) = .44$) than the latter
($S(\textit{paid}) = .66$), based on the probabilities of the meaning space
and the language. In sum, the model entertains several possible
interpretations upon encountering an ambiguous pronoun (some of which may be
preferred over others), and incrementally uses the incoming linguistic
information to resolve the pronoun and construct a coherent discourse-level
meaning interpretation. 

\subsection{Quantification}

Beyond assertive and negated sentences, the language $\mathcal{L}$
also contains existentially quantified (two-sentence) utterances in which a
quantified noun phrase (``someone'') is used as the subject (see
\cite{venhuizen2019framework} for a similar model that also includes
universally quantified utterances). The semantics for these utterances uses
existential quantification in meaning space, which results in a disjunctive
semantics over all possible subjects (i.e., the persons $p \in
\{\textit{elli}, \textit{mike}, \textit{nancy}, \textit{will}\}$). From its
linguistic experience, the model thus learns that ``someone'' can be used to
refer to any of these subjects, without preferring any particular
interpretation. Incremental navigation through meaning space, however, is
driven by both the linguistic experience of the model as well as the
structure of the meaning space. Given that the constituents of this
disjunctive semantics may themselves differ in terms of their probability in
meaning space, this will thus be reflected in the incremental inferences of
quantified utterances. Moreover, since all quantified expressions presented
to the model were part of a two-sentence utterance, the disjunctive
semantics was in some cases limited to either male or female persons,
depending on the pronoun used in the second sentence (``he/she'').

\begin{figure}
        \centering
        \includegraphics[width=\textwidth]{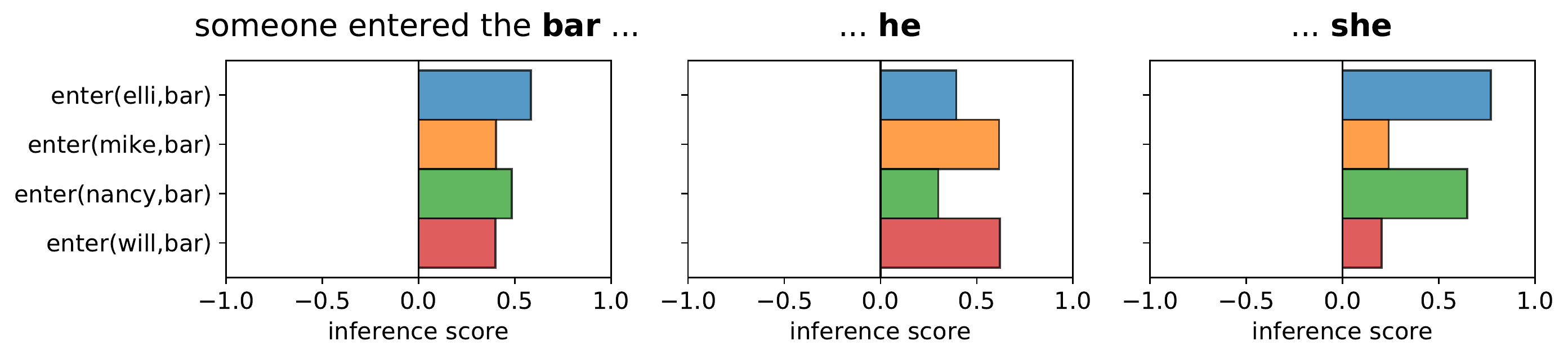}
        \caption{Processing of quantification. The barplots show the
        inference scores for a relevant set of propositions given the
        model's output for the word sequence ``someone entered the \textbf{bar}''
        (left) and the continuations ``someone entered the bar \textbf{he}''
        (middle) and ``someone entered the bar \textbf{she}''
        (right)---critical words are shown in boldface.}
        \label{fig:quantification}
\end{figure}

The interaction between quantification and anaphoric processing is
illustrated in Figure~\ref{fig:quantification}. The left-most panel shows a
selected set of relevant inferences for the sentence ``someone entered the
bar''. At the word ``bar'', the model shows moderately positive inferences
($\sim .5$) for each of the predicates \textit{enter(p,bar)}. The inferences
for \textit{enter(elli,bar)} (blue) and \textit{enter(nancy,bar)} (green)
are slightly higher than the others, which is explained by the fact that
they have a higher prior probability in the meaning space
($P(\textit{enter(elli,bar)}) = 0.29$; $P(\textit{enter(nancy,bar)}) =
0.25$; $P(\textit{enter(mike,bar)}) = 0.23$; $P(\textit{enter(will,bar)}) =
0.23$). The second and third panel show how these inferences change as the
sentence is continued using either a male (``he'') or a female pronoun
(``she''). As expected, the gender of the pronoun determines which
inferences are boosted; ``he'' results in higher inferences for
\textit{enter(mike,bar)} (orange) and \textit{enter(will,bar)} (red),
whereas ``she'' results in higher inferences for \textit{enter(elli,bar)}
and \textit{enter(nancy,bar)}. Moreover, the Surprisal estimates that derive
from the model again reflect the model's expectations about possible
continuations: the fact that the propositions describing females have a
slightly increased inference score already at the word ``bar'' results in
reduced Surprisal for ``she'' ($S(\textit{she}) = 0.92$) relative to ``he''
$S(\textit{he}) = 1.12$). 

\begin{figure}
        \centering
        \includegraphics[width=\textwidth]{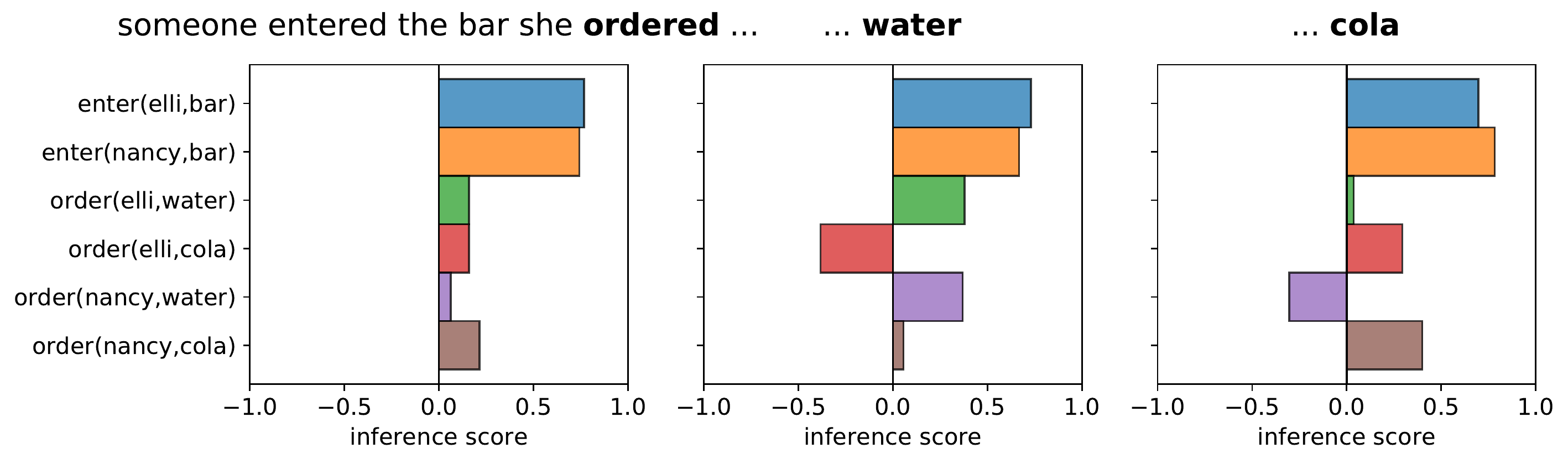}
        \caption{Processing of quantification (continued). The barplots show
        the inference scores for a relevant set of propositions given the
        model's output for the word sequence ``someone entered the bar she
        \textbf{ordered}'' (left) and the continuations ``someone entered
        the bar she ordered \textbf{cola}'' (middle) and ``someone entered
        the bar she ordered \textbf{water}'' (right)---critical words are
        shown in boldface.}
        \label{fig:inference}
\end{figure}

To further illustrate how the model's inferences incrementally develop in
the context of quantification, Figure~\ref{fig:inference} shows how
propositional inferences regarding \textit{nancy} and \textit{elli} develop
after processing the utterance ``someone entered the bar she ordered''. At
``ordered'', the model is in a state of indecision, in which
\textit{enter(elli,bar)} (blue) and \textit{enter(nancy,bar)} (orange) are
both positively inferred. Moreover, for \textit{elli} there is no preference
between ordering \textit{water} (green) and ordering \textit{cola} (red),
whereas for \textit{nancy} there seems to be a preference for \textit{cola}
(brown) as opposed to \textit{water} (purple)---this is in line with the
probabilistic constraints based on which the meaning space was created. When
the utterance-final word ``water'' is processed (middle panel), the model
maintains its disjunctive interpretation between \textit{enter(elli,bar)}
and \textit{enter(nancy,bar)} and moreover boosts its inferences for
\textit{order(elli,water)} and \textit{order(nancy,water)} to an equal
degree (while reducing the inferences for ordering \textit{cola}). Instead,
when ``cola'' is encountered as utterance-final word (right panel), the
model seems to show a slight preference for resolving its interpretation to
\textit{nancy}: \textit{enter(nancy,bar)} is boosted to a slightly higher
degree than \textit{enter(elli,bar)}, and \textit{order(nancy,cola)} obtains
a higher inference than \textit{order(elli,cola)}. This seems to suggest
that after processing ``cola'', the model is in a less uncertain state than
after processing ``water'', because there is a stronger preference for a
particular interpretation of the pronoun. The uncertainty of a point in
meaning space can be quantified using the notion of Entropy (see
Equation~\ref{eq:entropy}). Indeed, Entropy reveals that after ``cola'', the
model is in a slightly less uncertain state than after ``water''
($H(\vec{v}(\textit{water})) = 3.60$; $H(\vec{v}(\textit{cola})) = 3.51$).
Interestingly, the Surprisal estimates derived from the model seem to
suggest that---in this particular example---confirming the expectation for
\textit{order(nancy,cola)} is slightly less surprising ($S(\textit{cola}) =
0.52$) than disconfirming it ($S(\textit{water}) = 0.63$). For an
investigation of the interaction between meaning-level Entropy and Surprisal
during incremental meaning space navigation, see
\cite{venhuizen2019entropy}.

%% file: discussion.tex
\section{Discussion}\label{sec:discussion}

Distributional Formal Semantics defines propositional and sub-propositional
meaning within the meaning space, which is constituted from a set of
propositions $\mathcal{P}$, and a set of formal models $\mathcal{M}$. On the
propositional level, meaning vectors represent truth and falsehood with
respect to the set of models $\mathcal{M}$, thereby inherently capturing
co-occurrences between the propositions in $\mathcal{P}$. We have shown how
sub-propositional meaning within the meaning space derives from the
incremental navigation through meaning space, as modeled using a recurrent
neural network. Crucially, we showed how semantic phenomena such as
negation, presupposition, anaphoricity and quantification can be captured as
part of the semantic construction procedure and how they affect meaning
space navigation. Distributional Formal Semantics thus offers a powerful
synergy between formal and distributional approaches that paves the way
towards novel investigations into formal meaning representation and
construction. 

\subsection{DFS and distributional semantics offer complementary meaning representations}

Previous approaches that aimed to combine the strengths of formal and
distributional semantics have tried to do so by either expanding
distributional approaches to account for propositional-level inferences
\cite{coecke2011mathematical,baroni2014frege,
grefenstette2015concrete,rimell2016relpron}, or, conversely, by expanding
formal approaches with a distributional component to account for
lexical-level similarity
\cite{garrette2014formal,asher2016integrating,beltagy2016representing}. By
contrast, the DFS framework fundamentally integrates the distributional
hypothesis into a formal semantic model, while maintaining the
proposition-central perspective on meaning. That is, in terms of
propositional-level meaning, the distributional hypothesis can be formulated
as stating that propositions that occur in the same contexts---where
contexts are models that represent states of the world, rather than
linguistic contexts---represent similar meanings.

In DFS, the `representational currency' is propositions, whereas in
distributional semantics it is words. As a result, DFS allows us to model
similarity at the propositional level; e.g., $call(mike,waiter)$ is more
similar to $enter(mike,restaurant)$ than to $call(mike,bartender)$, since
the former two propositions tend to co-occur in $\mathcal{M}$, whereas
$call(mike,waiter)$ and $call(mike,bartender)$ never co-occur.
Distributional semantics, on the other hand, models lexical similarity in
terms of the distributional hypothesis \cite{Firth1957synopsis};
``\textit{waiter}'' is similar to ``\textit{bartender}'' as they occur in
similar linguistic contexts. Crucially, the DFS approach and distributional
semantics thus capture different notions of semantic similarity: while the
latter offers representations that inherently encode feature-based lexical
similarity between words, the former provides representations instantiating
truth-conditional similarity between propositions. Hence, we argue for a
division of labour between proposition-level and lexical distributional
representations in which the former captures meaning in terms of the state
of the world, and the latter captures the linguistic properties of lexical
items (see also \cite{Erk2016what}). In other words, we take the DFS
framework to be complementary to lexically-driven distributional semantics.

The complementary nature of these meaning representations is underlined by
recent advances in the neurocognition of language, where evidence suggests
that lexical retrieval (the mapping of words onto lexical semantics) and
semantic integration (the integration of word meaning into the unfolding
representation of propositional meaning) are two distinct processes involved
in word-by-word sentence processing (see
\cite{brouwer2017neurocomputational,brouwer2021neurobehavioral} for explicit
neurocognitive models). More specifically, this neurocognitive perspective
on comprehension suggests that there is no compositionality at the lexical
level; that is, word forms in context are mapped into word meaning
representations, which are subsequently integrated into a compositional
phrasal-/utterance-level meaning representation. Indeed, this suggests that
compositionality is only at play at the level of propositions, thus
eschewing the need for compositionality at the lexical level.

\subsection{Data-driven DFS}\label{sec:datadriven}

In DFS, the meaning of a proposition is captured by a meaning vector that
defines for each model $M \in \mathcal{M}$ that constitutes the meaning
space whether it satisfies the proposition or not. The quality of the
meaning representations therefore critically depends on finding a set of
models $\mathcal{M}$ that truth-conditionally and probabilistically capture
the structure of the world. In Section~\ref{sec:framework} we described an
algorithm to induce the meaning space from a high-level description of
the structure of the world. An advantage of this approach is that it allows
for defining the meaning space in a controlled manner, which is especially
important when modeling experimental data. From a computational perspective,
however, it would be interesting to investigate whether the meaning
space---or the underlying world knowledge constraints---could be empirically
derived from existing resources, such as knowledge bases and/or
linguistic corpora. 

There exists a wide variety of large-scale corpora that are annotated with
sentence-level semantic information (for an overview, see
\cite{bender2019linguistic}). For the purpose of generating a meaning space,
the annotations should minimally reflect predicate-argument structure (e.g.,
based on PropBank \cite{palmer2005propbank}, or FrameNet
\cite{baker1998framenet}). Crucially, however, a well-formed meaning space
captures \textit{all} co-occurrence probabilities between the propositions
that constitute the space. This means that the resource should not only
capture co-occurrence information between subsets of proposition-level
meanings (e.g, those that are part of a single discourse, as in the
Groningen Meaning Bank \cite{Bos2017gmb}, or appear in a single textual
entailment pair, as in the Stanford Natural Language Inference corpus
\cite{bowman2015large}), but rather across the entire set of
proposition-level meanings; that is, individual models in the DFS meaning
space describe the truth-values of all propositions in $\mathcal{P}$, which
would correspond to the union of the propositions occurring in all
discourses, or across all textual entailment pairs in a corpus. Moreover, in
order to maintain the distinction between probabilities deriving from the
structure of the world, and those deriving from linguistic experience (see
\cite{venhuizen2019expectation}), the co-occurrence probabilities
represented in the meaning space should ideally derive from
`world-knowledge'-driven inferences that are not confounded by linguistic
co-occurrence (although in many cases these will of course align).

An example of a resource that captures such `world knowledge'-driven
co-occurrences is the DeScript corpus \cite{wanzare2016descript}, which
contains crowd-sourced event sequence descriptions (ESDs) that describe
typical, everyday activities (also called `scripts'
\cite{schank1977scripts}). In addition, the DeScript corpus contains a gold
standard alignment for a subset of the crowd-sourced ESDs. Although the ESDs
are themselves linguistic, they contain `world-knowledge'-driven information
regarding the events and that are typically associated with a particular
script (e.g., baking a cake or fixing a bicycle), as well as the order in
which they typically occur. Sampling a meaning space from such a corpus
entails identifying a set of propositions $\mathcal{P}$ (in this case, the
events associated with a set of scripts), and collecting a large enough set
of models $\mathcal{M}$ that describe combinations between these
propositions; whereas using individual ESDs may result in data sparseness,
it is possible to sample models based on the co-occurrence information
derived from the individual ESDs. In future work, we plan to evaluate this
approach, and experiment with data-driven meaning space induction from
different resources.

\subsection{Set-theoretic meaning construction in DFS}\label{sec:settheory}

The DFS meaning space is constituted by proposition-level meanings that are
represented using binary meaning vectors. Since the meaning space itself is
continuous, however, sub-propositional meaning can be captured using
real-valued meaning vectors, which can be approximated using a Simple
Recurrent neural Network. Within the neural network model, the meaning of a
sub-propositional expression can be interpreted as the point in meaning
space that is in between the (propositional) meaning vectors that represent
the meanings of all possible continuations (see Figure~\ref{figure:cube}).
In other words, the model navigates to a point in meaning space that is
consistent with all interpretations that are captured by the current
sub-propositional expression. An alternative approach to defining
sub-propositional meaning is using sets of proposition-level meaning vectors
(cf. \cite{baroni2010nouns}). That is, in line with the generalized
quantifier approach that represents the meaning of a quantified expression
as a set of propositions, the meaning of a sub-propositional expression in
the meaning space can be represented as the set of propositional meaning
vectors consistent with its meaning. For instance, the meaning of the proper
name ``mike'' will be captured by the set of meaning vectors that describe
the propositions that pertain to \textit{mike}: $\llbracket \text{mike}
\rrbracket_{\mathcal{S}_{\mathcal{M}\times\mathcal{P}}} = \{\vec{v}(p) | p
\in \{\textit{enter(mike,bar)}, \textit{enter(mike,restaurant)},
\textit{call(mike,bartender)}, \ldots\}\}$. Since each meaning vector
$\vec{v}(p)$ can be interpreted as describing the set of models that satisfy
$p$, this set-theoretic interpretation effectively describes
sub-propositional expressions as sets of sets of models. 

Meaning construction, then, entails defining operations on sets of meaning
vectors. In particular, the disjuction between two sets of meaning vectors
can simply be defined as the union between these sets. Conjunction between
sets of meaning vectors, in turn, means conjoining all elements of the first
set of meaning vectors with all elements of the second set (based on the
composition operation defined in Section~\ref{sec:composition}). Since
negating a set of meaning vectors implies that none of these is the case,
the negation of a set of meaning vectors can be defined as the conjunctive
closure over the negations of all individual meaning vectors. In order to
capture the incrementality of meaning construction, a `merge' operation (cf.
merge between Discourse Representation Structures in DRT
\cite{kamp1993discourse}) can be defined that asserts a set of propositional
meaning vectors $V$ into a context $C$---which is also defined as a set of
meaning vectors---by selecting the subset of meaning vectors in $V$ that is
consistent with context $C$. A full description of set-theoretic
interpretation in DFS is beyond the scope of the current manuscript, but an
initial formalization of these operations can be found as part of
\textsc{dfs-tools}. 

Crucially, sets of meaning vectors can be mapped back into the meaning space
by finding the point in meaning space that is in between all elements of a
given set. Formally, this means that the point in meaning space that
corresponds to a set of meaning vectors is the (real-valued) vector that
constitutes the arithmetic mean between these meaning vectors. In fact,
given a completely balanced linguistic input, the recurrent neural network model is
predicted to approximate exactly these intermediate points in space when
capturing sub-propositional meaning. Indeed, it is the interaction between
the structure of the meaning space and the structure of the linguistic
experience that is difficult (if at all possible) to capture in an `offline'
set-theoretic approach, while it is an inherent part of the way in which the
neural network model maps language onto meaning.

\subsection{DFS in cognitive models of language processing}

DFS offers a powerful and flexible framework for modeling meaning and
probabilistic inference in cognitive models of human language processing.
For instance, a neural network model of language comprehension, similar to
the one presented above, but employing meaning representations derived from
an earlier formulation of the DFS framework (see \cite{Frank2003modeling}),
has been used to successfully model the interaction between linguistic
experience and world knowledge in comprehension
\cite{venhuizen2019expectation}. Moreover, models employing such meaning
representations have been shown to naturally capture inference and
quantification \cite{Frank2009connectionist}, and generalize to unseen
sentences and semantics, in both comprehension \cite{Frank2009connectionist}
and production \cite{calvillo2016connectionist}.  Here, we have extended
these results by showing how they capture phenomena such as negation,
presupposition, and anaphoricity. Finally, as discussed above, the
information-theoretic notions of Surprisal and Entropy directly derive from
DFS representations (see
\cite{venhuizen2019expectation,venhuizen2019entropy}), thereby providing
representation-grounded linking hypotheses between processing behavior in
the model and behavioral correlates of human processing difficulty.
Crucially, this spectrum of linking hypotheses can be extended even further
by employing DFS representations in a neurocomputational model of the
electrophysiology of language comprehension to also obtain direct estimates
of neurophysiological processing, in particular, of the N400 and P600
components of the Event-Related brain Potential (ERP) signal; see
\cite{brouwer2021neurobehavioral}. 

Taken together, we believe that the power and flexibility of DFS
representations, the incremental construction of these representations in
neural network models, and the representation-grounded linking hypotheses to
behavioral and neurophysiological indices of processing difficulty, provide
a comprehensive workbench for 1) formal semantic theory grounded in
psycholinguistic evidence, 2) offering foundations for psycholinguistic
theory by formally explicating representations and mechanisms, and 3)
overall integration of formal semantic and psycholinguistic approaches to
the study of language.  As such, we believe DFS has the potential to shed
light on the processing nature, as well as the representational and
mechanistic underpinnings thereof, of a vast spectrum of syntactic,
semantic, and pragmatic phenomena.

%% file: conclusion.tex
\section{Conclusion}

We have proposed a framework for Distributional Formal Semantics (DFS), in
which (sub-)propositional meaning is defined relative to a meaning space,
which is constituted by a set of first-order models
$\mathcal{M}_\mathcal{P}$ that is defined relative to a set of propositions
$\mathcal{P}$ (such that each model $M \in \mathcal{M}_\mathcal{P}$ can be
defined in terms of the subset of propositions $P \subseteq \mathcal{P}$
that it satisfies). Within this meaning space, propositional meaning is
represented by the meaning vector $\vec{v}(p)$ that represents the meaning
of proposition $p \in \mathcal{P}$ in terms of the models that satisfy $p$;
i.e., $\vec{v}(p)$ is the vector that assigns $1$ to all
$M\in\mathcal{M}_\mathcal{P}$ such that $p$ is satisfied in $M$. The meaning
vector $\vec{v}(p)$ thus defines the meaning of $p$ as a point in the
meaning space. Within the meaning space, similar meanings (e.g.,
propositions with high co-occurrence) obtain similar meaning vectors that
are positioned close to each other in space.

We have shown that the resultant distributed meaning representations are
inherently compositional and probabilistic, and that they allow for
capturing probabilistic inference and entailment. Moreover, we have shown
how the information-theoretic notions of Surprisal (the `self-information'
of a transition in meaning space) and Entropy (the average Surprisal over
all possible transitions from one point to the next) derive from these
representations. Furthermore, sub-propositional meanings, which cannot be
directly expressed as (combinations of) propositions, can be represented as
real-valued vectors, constituting points in the meaning space that also
capture their own probability. To derive these sub-propositional
representations, we instantiated a semantic interpretation function that
maps utterances onto DFS meaning representations. Rather than formalizing
such a function using set-theoretic machinery, we have employed a recurrent
neural network model to incrementally construct the semantics for an
utterance by navigating the meaning space on a word-by-word basis.
Crucially, we have shown that this model of incremental semantic
construction naturally captures semantic phenomena such as negation,
quantification, presupposition and anaphoricity, and how these phenomena
affect the incremental processing dynamics of the model, as quantified by
probabilistic inference, Surprisal, and Entropy.

DFS integrates the strengths of formal semantics and distributional
semantics by incorporating a distributional component into a formal system.
In DFS, like in formal semantics, the representational currency is
propositions. In contrast to distributional semantics, which defines lexical
meaning in terms of word co-occurrences, meaning in DFS is defined in terms
of propositional co-occurrence, which reflects hard and probabilistic world
knowledge constraints. As a result, we take DFS to be complimentary to
distributional semantics; that is, where distributional semantics offers the
representational machinery to capture lexical meaning, DFS offers
utterance-level meaning representations. This distinction is in line with
recent psycholinguistic theorizing, which suggests that compositionality may
only be required at the utterance-level, thereby calling into question the
need to directly incorporate aspects from formal semantics into
distributional semantics.

In sum, DFS offers a powerful synthesis between formal and
distributional approaches to semantics: distributed, utterance-level meaning
representations that capture the similarity between (propositional)
meanings, while strictly maintaining the formal properties of
compositionality and entailment. As such, we believe that the DFS
framework---implemented by \textsc{dfs-tools}---paves the way towards novel
investigations into the representation and construction of utterance-level
meanings, and the relation between the formal, empirical and cognitive study
of semantics.

%% file: appendix.tex
\section{Probabilistic constraints}
\label{app:prob_constraints}

The inference-based sampling algorithm employs probabilistic constraints to
determine the truth/falsehood of propositions that cannot be inferred, which
is the case if a proposition is consistent both with respect to the Light
World and with respect to the Dark World (see Algorithm~\ref{alg:sampling}
in Section~\ref{sec:sampling}). The set of probabilistic constraints used
for sampling our meaning space is shown in Table~\ref{tab:prob_constraints}.
During sampling, the truth of $p$ is probabilistically determined relative
to model $M$---the Light World---based on the probability described by
$Pr(p,M)$, given that $M$ satisfies pre-defined conditions (see the third
column).  The sampling algorithm employs these probabilistic constraints in
an ordered manner such that the first matching constraint determines the
probability. This means that if the model $M$ that describes the
state-of-affairs sampled so far does not satisfy any of the conditions
defined for proposition $p$, $p$ is sampled according to the base sampling
probability (see constraint 19).\footnote{The base sampling probability for
propositions (see constraint 19) is set to $0.6$ instead of a coin flip
($0.5$) in order to increase propositional co-occurrence across models.}
Note that a sampling probability of $1$ (as in probabilistic constraints
14-17) means that whenever the truth of $p$ is determined probabilistically,
that is, if the truth/falsehood of $p$ cannot be inferred, it will be
assigned the truth value `true' (with respect to the Light World).
Conversely, a sampling probability of $0$ (as in probabilistic constraint
$18$) means that $p$ will never be probabilistically determined to be
`true'---i.e., $p$ will only be true relative to the Light World due to
direct inference based on the hard world knowledge constraints (as described
in Section~\ref{sec:constructing}). Hence, due to the inference-based nature
of the sampling algorithm, the observed probabilities in the sampled meaning
space may only indirectly reflect the sampling probabilities shown in
Table~\ref{tab:prob_constraints} (see Section~\ref{sec:constructing} for
discussion).

\addtocounter{footnote}{-1} 
\begin{table}
        \caption{Probabilistic constraints used for sampling of
        propositions for the meaning space. The first column identifies the
        order and the second column provides a description of each of the
        probabilistic constraints. The sampling probability ($Pr(X,M)$),
        shown in column four, describes the sampling probability of proposition
        $X$ relative to model $M$, given that $M$ satisfies the conditions
        presented in column three. Individual variables correspond to
        logical constants of a particular type: persons ($p \in
        \{\textit{mike, will, elli, nancy}\}$), places ($l \in
        \{\textit{bar, restaurant}\}$), servers ($s \in \{\textit{barman,
        waiter}\}$), and orders ($o_{\textit{food}} \in
        \{\textit{fries,salad}\}$; $o_{\textit{drink}} \in \{\textit{cola,
        water}\}$). The variables $x$ and $y$ are of the general type of
        entities, and $X$ is of the type of propositions.}
        \label{tab:prob_constraints}
        \small
        \begin{tabularx}{\textwidth}{c X l l}
                \toprule
                \textbf{No.} 
                & \textbf{Probabilistic Constraint}
                & \textbf{Model Conditions}
                & \textbf{Sampling probability} \\
                \midrule
                1 
                & Persons tend to enter the same place
                & $M\vDash \exists x.\textit{enter(x,l)}$
                & $Pr(\textit{enter(p,l)},M) = 0.9$ \\
                2 
                & Entering bar unlikely if person orders food
                & $M\vDash\exists o_{\textit{food}}.\textit{order}(p,o_{\textit{food}})$
                & $Pr(\textit{enter(p,bar)},M) = 0.1$ \\
                3 
                & Entering bar likely if person orders drink
                & $M\vDash\exists o_{\textit{drink}}.\textit{order}(p,o_{\textit{drink}})$
                & $Pr(\textit{enter(p,bar)},M) = 0.9$ \\
                4  
                & Entering rest. likely if person orders food
                & $M\vDash\exists o_{\textit{food}}.\textit{order}(p,o_{\textit{food}})$
                & $Pr(\textit{enter(p,restaurant)},M) = 0.9$ \\
                5  
                & Low probability of ordering food in bar
                & $M\vDash\textit{enter(p,bar)}$
                & $Pr(\textit{order}(p,o_{\textit{food}}),M) = 0.1$ \\
                6  
                & High probability of ordering food in restaurant
                & $M\vDash\textit{enter(p,restaurant)}$
                & $Pr(\textit{order}(p,o_{\textit{food}}),M) = 0.9$ \\
                7  
                & Different persons unlikely to order the same order
                & $M\vDash\exists x.x\neq p \wedge \textit{order}(x,o)$
                & $Pr(\textit{order}(p,o),M) = 0.1$\\
                8 
                & Order unlikely to be brought if not ordered
                & $M\vDash \neg \exists x.\textit{order}(x,o)$ 
                & $Pr(\textit{bring}(s,o),M) = 0.1$\\
                9 
                & Low probability of paying if someone else pays in the
                same place
                & \multirow{2}{3.6cm}{$M\vDash \exists x\exists y.\textit{enter}(p,x) \wedge y\neq p \wedge \textit{enter}(y,x) \wedge \textit{pay}(y)$}
                & $Pr(\textit{pay}(p),M) = 0.1$\\
                10  
                & Personal preference (\textit{elli})
                & $M \vDash \top$
                & $Pr(\textit{order(elli,cola)},M) = 0.9$\\
                11 
                & Personal preference (\textit{mike})
                & $M \vDash \top$
                & $Pr(\textit{order(mike,cola)},M) = 0.9$\\
                12 
                & Personal preference (\textit{nancy})
                & $M \vDash \top$
                & $Pr(\textit{order(nancy,water)},M) = 0.9$\\
                13 
                & Personal preference (\textit{will})
                & $M \vDash \top$
                & $Pr(\textit{order(will,water)},M) = 0.9$\\
                14  
                & Bar presupposes bartender
                & $M\vDash \textit{referent(bar)}$
                & $Pr(\textit{referent(bartender)},M) = 1$ \\
                15 
                & Restaurant presupposes waiter
                & $M\vDash \textit{referent(restaurant)}$ 
                & $Pr(\textit{referent(waiter)},M) = 1$ \\
                16  
                & Bartender presupposes bar
                & $M\vDash \textit{referent(bartender)}$
                & $Pr(\textit{referent(bar)},M) = 1$\\
                17  
                & Waiter presupposes restaurant
                & $M\vDash \textit{referent(waiter)}$
                & $Pr(\textit{referent(restaurant)},M) = 1$\\
                18 
                & Base probability referents (infer only)
                & $M \vDash \top$
                & $Pr(\textit{referent(x)},M) = 0$\\
                19 
                & Base probability propositions\footnotemark
                & $M \vDash \top$
                & $Pr(X,M) = 0.6$\\
                \bottomrule
            \end{tabularx}
        \end{table}

        \begin{algorithm}[t]
                \caption{Model selection algorithm that reduces a set of models
                $\mathcal{M_P}$ to a subset $\mathcal{M}^{*}_\mathcal{P}$ of
                cardinality $k$, while preserving the structure of $\mathcal{M_P}$
                truth-conditionally and probabilistically.}
                \label{alg:selection}
        \begin{enumerate}\small
                \item Take a random proper subset of $k$ models from
                $\mathcal{M}_\mathcal{P}$, and call this $\mathcal{M}^{*}_\mathcal{P}$;
        
                \item Check if all propositional meaning vectors in
                $\mathcal{M}^{*}_\mathcal{P}$ are informative (i.e., there are no
                vectors that only contain zeros), otherwise return to step (1);
        
                \item Compute an inference vector $\vec{\textit{inf}}$ containing
                the inference score $\textit{inference}(a,b)$ for each combination
                of propositions $a,b \in \mathcal{P}$ for the original set of models
                ($\vec{\textit{inf}}(\mathcal{M}_\mathcal{P})$) and the reduced set
                of models ($\vec{\textit{inf}}(\mathcal{M}^{*}_\mathcal{P})$).
        
                \item Check if the reduced set of models encodes the same hard
                constraints as the original set of models (positive constraints: for
                all $i$, $\vec{\textit{inf}}(\mathcal{M}^{*}_\mathcal{P})(i) = 1$
                \textit{iff} $\vec{\textit{inf}}(\mathcal{M}_\mathcal{P})(i) = 1$;
                negative constraints: for all $i$,
                $\vec{\textit{inf}}(\mathcal{M}^{*}_\mathcal{P})(i) = -1$
                \textit{iff} $\vec{\textit{inf}}(\mathcal{M}_\mathcal{P})(i) = -1$),
                otherwise return to step (1);
                
                \item Compute the similarity between $\mathcal{M}^{*}_\mathcal{P}$
                and $\mathcal{M}_\mathcal{P}$ on the basis of the
                proposition-by-proposition inference scores in
                $\vec{\textit{inf}}(\mathcal{M}^{*}_\mathcal{P})$ and
                $\vec{\textit{inf}}(\mathcal{M}_\mathcal{P})$ using Pearson's
                correlation coefficient
                $\rho(\vec{\textit{inf}}(\mathcal{M}^{*}_\mathcal{P}),\vec{\textit{inf}}(\mathcal{M}_\mathcal{P}))$;
                
                \item If $\mathcal{M}^{*}_\mathcal{P}$ is the best approximation of
                $\mathcal{M}_\mathcal{P}$ so far ($\rho > \rho_{\textit{best}}$),
                store it;
                
                \item Start from step (1) and run the next iteration;
                
                \item If we have reached $X$ iterations, return the best
                $\mathcal{M}^{*}_\mathcal{P}$ found.
        \end{enumerate}
        \end{algorithm}
        
\section{Model Selection algorithm}\label{app:selection}

We employ a model selection algorithm to reduce the dimensionality of the
meaning space (following \cite{venhuizen2019expectation}). This means that
based on a given set of models $\mathcal{M}_\mathcal{P}$, we define a new
set of models $\mathcal{M}^{*}_\mathcal{P}$ such that: (i)
$\mathcal{M}^{*}_\mathcal{P} \subset \mathcal{M}_\mathcal{P}$; (ii)
$\mathcal{M}^{*}_\mathcal{P}$ captures the same hard world knowledge
constraints as $\mathcal{M}_\mathcal{P}$; and (iii)
$\mathcal{M}^{*}_\mathcal{P}$ approximates the probabilistic inferences
derived from $\mathcal{M}_\mathcal{P}$.
Formally, we reduce the dimensionality of a meaning space by reducing the
number of models in $\mathcal{M}_\mathcal{P}$ to $k$, such that $k <
|\mathcal{M}_\mathcal{P}|$. The procedure, shown in
Algorithm~\ref{alg:selection}, is repeated for $X$ iterations to arrive at a
set of models $\mathcal{M}^{*}_\mathcal{P}$ (with
$|\mathcal{M}^{*}_\mathcal{P}| = k$) that maximally reflects the knowledge
encoded in the original space $\mathcal{M}_\mathcal{P}$. For the present
model, we selected $150$ models from the sampled set of 10K models (based on
$50$ iterations), resulting in a reduced meaning space with
$\rho(\vec{\textit{inf}}(\mathcal{M}^{*}_\mathcal{P}),\vec{\textit{inf}}(\mathcal{M}_\mathcal{P}))
= .91$.